\documentclass[10pt,twocolumn,letterpaper]{article}

\usepackage[pagenumbers]{wacv} %

\usepackage{dsfont}
\usepackage{etoolbox}

\newif\ifshowedits

\newcommand{\addeditor}[3]{%
  \definecolor{#1color}{rgb}{#3}
  \expandafter\newcommand\csname #1\endcsname[1]{%
  \ifshowedits
    {\color{#1color} ##1}%
  \else
    {##1}%
  \fi
  }%
  \expandafter\newcommand\csname #1rmk\endcsname[1]{%
  \ifshowedits
    {\color{#1color} {\bf [#2: ##1]}}
  \fi
  }%
  \expandafter\newcommand\csname #1rpl\endcsname[2]{%
  \ifshowedits
    {\color{#1color} ##1 \sout{##2}}
  \else
    {##1}
  \fi
  }%
}

\newcommand{\createtextvar}[1]{
  \expandafter\newcommand\csname #1\endcsname{%
  {\text{#1}}
}%
}

\newcommand{\mycomment}[1]{}

\newcommand{\calD}{{\cal D}}

\newcommand{\calX}{{\cal X}}
\newcommand{\calY}{{\cal Y}}

\newcommand{\bx}{{\bf x}}

\usepackage[table,xcdraw]{xcolor}
\usepackage{multirow} %
\usepackage{booktabs}       %
\usepackage{colortbl}       %
\usepackage{pifont} %
\usepackage{tikz}
\usetikzlibrary{shapes.geometric}
\usepackage{pgfplots}
\usetikzlibrary{plotmarks}

\definecolor{baseline}{HTML}{000000}          %
\definecolor{advtrain}{HTML}{FF7F0E}          %
\definecolor{augment}{HTML}{2CA02C}           %
\definecolor{contrastive}{HTML}{E41A1C}       %
\definecolor{recipes}{HTML}{9467BD}           %
\definecolor{freezing}{HTML}{8C564B}          %

\definecolor{classification}{HTML}{1f77b4}   %
\definecolor{feature}{HTML}{17becf}          %
\definecolor{hybrid}{HTML}{bcbd22}           %
\definecolor{inter}{HTML}{e377c2}            %
\definecolor{gradients}{HTML}{7f7f7f}           %

\DeclareRobustCommand{\mycircle}[1]{%
  \tikz[baseline=-0.6ex]\fill[#1] (0,0) circle (0.6ex);%
}
\definecolor{mygreen}{rgb}{0.17,0.63,0.17}

\newcommand{\nmodels}{56\xspace} 
\newcommand{\nmethods}{21\xspace} 
\newcommand{\noodsets}{eight\xspace} 
\newcommand{\noodcategories}{four\xspace}

\definecolor{wacvblue}{rgb}{0.21,0.49,0.74}
\usepackage[pagebackref,breaklinks,colorlinks,allcolors=wacvblue]{hyperref}

\def\confName{WACV}
\def\confYear{2026}

\title{One Model, Many Behaviors: Training-Induced Effects on Out-of-Distribution Detection}

\author{
Gerhard Krumpl\thanks{Correspondence: \tt gerhard.krumpl@icg.tugraz.at} \textsuperscript{$\,$1,2} \qquad
Henning Avenhaus\textsuperscript{2} \qquad 
Horst Possegger\textsuperscript{1} \\ [0.4em]
\textsuperscript{1}Institute of Visual Computing, Graz University of Technology, Austria\\
\textsuperscript{2}KESTRELEYE GmbH, Austria\\
}

\begin{document}
\maketitle
\begin{abstract}

Out-of-distribution (OOD) detection is crucial for deploying robust and reliable machine-learning systems in open-world settings. 
Despite steady advances in OOD detectors, their interplay with modern training pipelines that maximize in‑distribution (ID) accuracy and generalization remains under-explored.
We investigate this link through a comprehensive empirical study.
Fixing the architecture to the widely adopted ResNet‑50, we benchmark 21 post-hoc, state-of-the-art OOD detection methods across 56 ImageNet-trained models obtained via diverse training strategies and evaluate them on eight OOD test sets.
Contrary to the common assumption that higher ID accuracy implies better OOD detection performance, we uncover a non‑monotonic relationship: OOD performance initially improves with accuracy but declines once advanced training recipes push accuracy beyond the baseline.
Moreover, we observe a strong interdependence between training strategy, detector choice, and resulting OOD performance, indicating that no single method is universally optimal.

\end{abstract}
    
\section{Introduction}
\label{sec:intro}

\begin{figure}[t]
  \centering
   \includegraphics[width=1.0\linewidth]{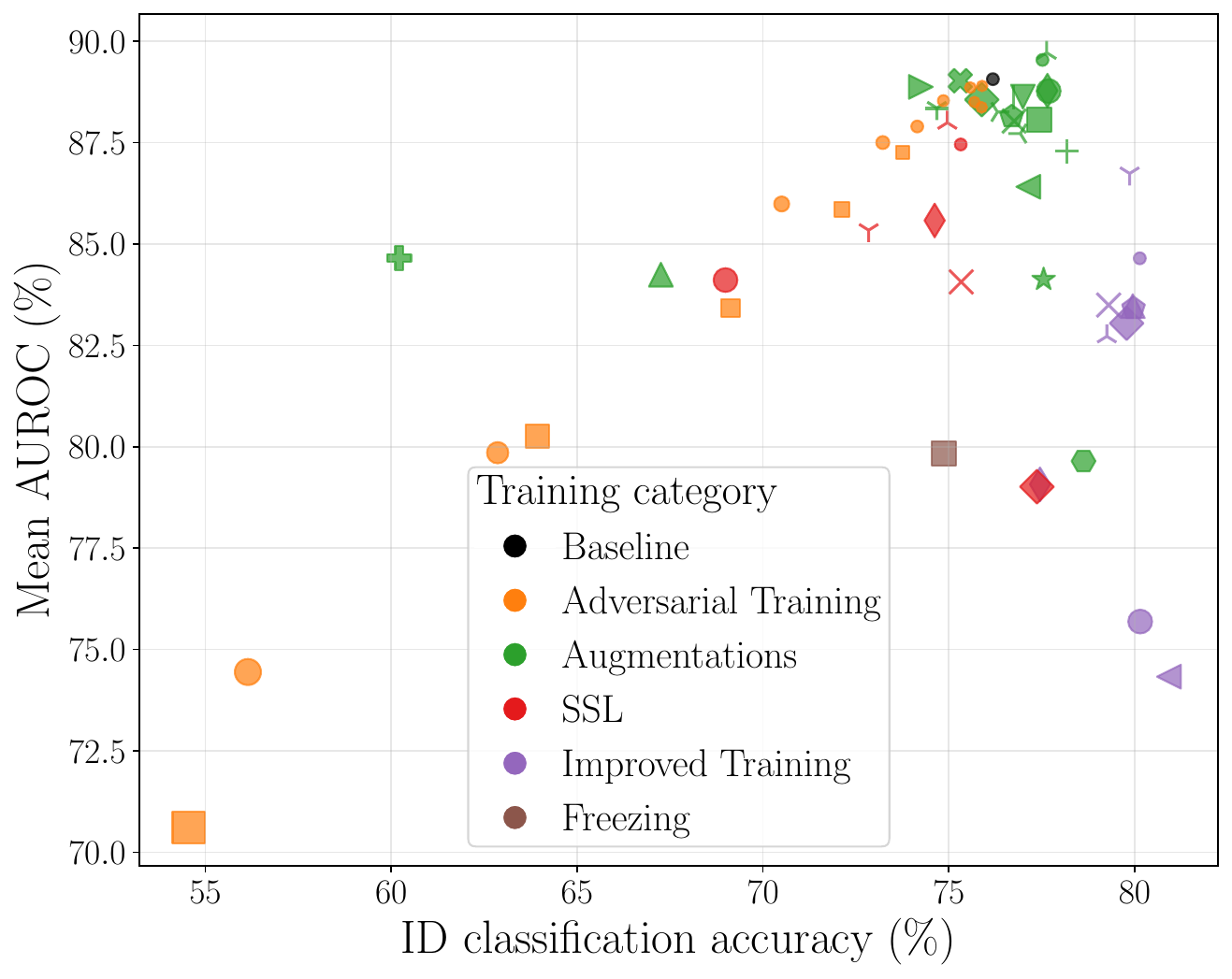}

   \caption{
        \textbf{Top-1 classification accuracy is an unreliable indicator for OOD detection performance.}
        This figure shows the relationship between in-distribution (ID) classification accuracy and out-of-distribution (OOD) detection performance for \nmodels{} ResNet-50 models, all sharing the same architecture but trained with different strategies.  
        Each point represents one model and reports the mean AUROC (Area Under the Receiver Operating Characteristic Curve) over \nmethods{} OOD detection methods and \noodsets{} OOD datasets.
        Color indicates the model's training category, while the marker shape uniquely identifies each model within that category.  
   }
   \label{fig:teaser}
\end{figure}

The robust deployment of machine learning (ML) systems in real-world environments depends on models that not only perform well on in-distribution (ID) data but can also reliably detect inputs that differ from their training distribution---such as novel object categories, unexpected environmental conditions, or sensor corruptions.
When exposed to such out-of-distribution (OOD) inputs, modern ML models often produce high-confidence yet incorrect predictions~\cite{Nguyen2014, moosavidezfooli17, hein2019relu}. 
This behavior poses significant risks in safety-critical settings, such as autonomous driving, healthcare, or industrial quality assurance. 
For example, an autonomous car might find itself in an unexpected emergency scene, or a food-sorting stream could encounter unexpected (potentially allergy-inducing) foreign materials.

Post-hoc OOD detection methods offer a promising solution. %
These methods are appealing in practice: i) they are simple to deploy on pretrained models, avoiding costly retraining and preserving ID accuracy, and ii) they require no prior exposure to OOD data, which, by definition, are unknown before deployment. 
A wide variety of methods have emerged, ranging from simple confidence scores~\cite{hendrycks2016_ood_msp} to more sophisticated approaches based on feature-space distances~\cite{lee2018_mahala, sun2022knnood, Ren2021_rmds} and model enhancements~\cite{sun2021_ood_react, djurisic2023ash}.

Concurrently, modern classification training pipelines have become increasingly diverse and powerful. 
Augmentations such as MixUp~\cite{zhang2018mixup} or CutMix~\cite{Sangdoo2019_cutmix}, regularization techniques like label smoothing~\cite{Szegedy_label_smooting} or Exponential Moving Averages (EMA), contrastive/self-supervised pretraining, and extended training schedules have all been shown to improve ID accuracy and generalization~\cite{wightman2021resnet, He2018BagOT, Vasilis2023_TV2}. 
Yet, surprisingly, the impact of such training strategies on post-hoc OOD detection has received little attention. 
Most prior evaluations continue to benchmark OOD detection methods on models trained with vanilla strategies---failing to reflect the diversity of models used in practice.

A widely held assumption is that higher ID accuracy naturally translates into stronger OOD detection~\cite{vaze2022openset, galil2023fwbm, Humblot-Renaux_2024_CVPR}. 
This belief has shaped experimental design and practical implementation alike: train a more accurate classifier and get stronger OOD detection performance \textit{for free}. 
However, this assumption is rarely questioned---and its validity under diverse training strategies remains unclear.

In this work, we revisit this and other assumptions through the first large-scale, architecture- and data-controlled study on the effects of training strategies on post-hoc OOD detection. 
By fixing the architecture (ResNet-50~\cite{He2015_resnet}) and dataset (ImageNet~\cite{deng2009_imagenet}), we isolate the impact of training strategy alone. 
We evaluate \nmethods post-hoc OOD detection methods across \nmodels models spanning four training families: data augmentation, contrastive/self-supervised learning, adversarial training, and improved training schedules---as summarized in \cref{fig:teaser}. 
Performance is assessed using eight benchmark OOD datasets that cover near-, far-, extreme-, and synthetic-OOD categories.

This study provides new insights into how training influences OOD detection and offers guidance for developing more robust, generalizable OOD detection methods.
We summarize our contribution as follows:
\begin{itemize}
    \item 
    We conduct a large-scale, architecture- and data-controlled study on the influence of training strategies on post-hoc OOD detection, covering \nmodels models, \nmethods OOD detectors, and \noodsets OOD test sets. 
    We further analyze the correlation between ID accuracy and OOD detection performance across methods and training strategies.

    \item 
    We show that the commonly assumed correlation between ID accuracy and OOD robustness is not generally valid. 
    Prior studies often focus on low-to-moderate accuracy regimes or vary model architectures, where capacity effects confound the relationship.

    \item
    We identify substantial variance in method robustness: many post-hoc detectors overfit to vanilla training recipes and degrade under diverse strategies, while methods that utilize richer internal representations generalize better.
\end{itemize}

\section{Related Work}
\label{sec:related_work}
Out-of-distribution (OOD) detection methods are broadly categorized into training-based and post-hoc approaches~\cite{yang2021_ood_survey}.
Training-based methods modify model architectures, loss functions, or regularization strategies to improve OOD robustness~\cite{lin2021mood, du2022_vos, Hsu2020_godin}. 
Recent training-based works also focus on improving OOD generalization and detection~\cite{bai2023_feed_two_birds, zhu2024_croft}.
In contrast, post-hoc methods operate on pretrained models without requiring retraining or access to OOD samples.
We focus on post-hoc, sample-free methods due to their practical advantages: they are model-agnostic, preserve the model's in-distribution (ID) performance, and scale efficiently to large deployments, where retraining is costly or infeasible~\cite{zhang2024openood}.

Post-hoc methods can be further grouped into score-based and model enhancement methods.
\emph{Score-based methods} assign an OOD score to each input based on model outputs or internal activations.
Within this category, classification-based methods were pioneered by Hendrycks \etal~\cite{hendrycks2016_ood_msp}, who introduced Maximum Softmax Probability (MSP), which uses the highest softmax output as a baseline OOD score.
Subsequent variants, such as Maximum Logit Score (MLS)~\cite{hendrycks2019_ood_mls} and Energy-Based OOD (EBO)~\cite{liu2020_ood_ebo}, refine this approach by operating directly on the model’s logits or their transformations.
ODIN~\cite{shiyu17} further improves separation by combining temperature scaling with input perturbations.
Feature-based methods instead operate on internal representations: Mahalanobis distance score (MDS)~\cite{lee2018_mahala}, Simplified Hopfield Energy (SHE)~\cite{zhang2023_she}, and KNN-based methods~\cite{sun2022knnood} compute distances in feature space, while GRAM~\cite{sastry20a_gram} and GradNorm~\cite{huang2021_gradnorm} leverage intermediate statistics or input gradients.
\emph{Model enhancement} methods modify inference-time activations to improve OOD detection robustness without changing model weights.
ReAct~\cite{sun2021_ood_react} improves robustness by clipping abnormal activations, ASH~\cite{djurisic2023ash} prunes and rescales feature activations to down-weight irrelevant neurons, and SCALE~\cite{xu2024scaling} shapes internal representations.
These are typically paired with energy-based scoring and have demonstrated strong performance in large-scale benchmarks~\cite{zhang2024openood}.

\paragraph{ID accuracy vs.~OOD detection performance.}  
Prior works~\cite{vaze2022openset, galil2023fwbm, Humblot-Renaux_2024_CVPR} have reported a positive correlation between ID accuracy and OOD detection performance.
Vaze \etal~\cite{vaze2022openset} found a significant correlation between closed-set classification accuracy and open-set recognition performance. 
Similarly, Galil \etal~\cite{galil2023fwbm} noted this correlation across various pre-trained deep neural networks when using Maximum Softmax Probability (MSP) as the OOD score.
Humbold-Renaux \etal~\cite{Humblot-Renaux_2024_CVPR} also identified a connection between ID and OOD performance, specifically examining the impact of label noise on OOD detection within a small-scale benchmark.
More recently, Wang \etal~\cite{wang2024dissect} extended this line of inquiry by analyzing the distinction between closed- and open-set recognition, highlighting how auxiliary OOD data influence performance. 
Notably, they reported a negative correlation between closed- and open-set recognition in large-scale settings.
Complementary to these findings, in the field of OOD generalization, OoD-Bench \cite{ye2022_oodbench} studies diversity and correlation shifts under a fixed label space (no label shift), showing how models that learn spurious cues (\eg, in Colored MNIST \cite{arjovsky2020invariantriskminimization}) achieve high accuracy without robustness to distribution shifts.

The works in~\cite{vaze2022openset, galil2023fwbm} evaluate a wide variety of pretrained deep neural networks, spanning various architectures and focusing primarily on MSP or a limited subset of baseline OOD detection methods.
Similarly, \cite{Humblot-Renaux_2024_CVPR} investigates the impact of label noise while varying both ID datasets and model architectures.
Varying multiple factors simultaneously, such as architecture, model capacity, and ID dataset, introduces potential confounding effects, making it difficult to disentangle the individual impact of training-related factors.

To address this, we take a more controlled experimental design: we fix the model architecture and the ID dataset. 
This allows us to isolate the impact of different training strategies and to reevaluate the relationship between ID accuracy and OOD detection performance across a wide range of post-hoc methods and diverse OOD test distributions.

\section{Investigating Training-induced Effects}
\label{sec:methodology}
This section addresses our central question: How do different training strategies influence the effectiveness of post-hoc OOD detectors?
While prior studies~\cite{zhang2024openood,vaze2022openset,Humblot-Renaux_2024_CVPR} often vary the architecture or in-distribution (ID) dataset, which confounds the effects of training, we isolate the impact of the training strategy by fixing both: a ResNet-50 architecture trained on ImageNet.

Our study evaluates \nmethods{} sample-free post-hoc OOD detection methods on \nmodels{} ResNet-50 models, each following a distinct training strategy.%
Evaluation covers \noodsets OOD test datasets grouped into four categories: near-, far-, extreme-, and synthetic-OOD.
This design enables a systematic investigation of how training choices affect OOD detection. %

The following subsections detail the problem setting, datasets, models, metrics, and detection methods that underpin our study.

\begin{figure*}
    \centering
    \includegraphics[width=\linewidth]{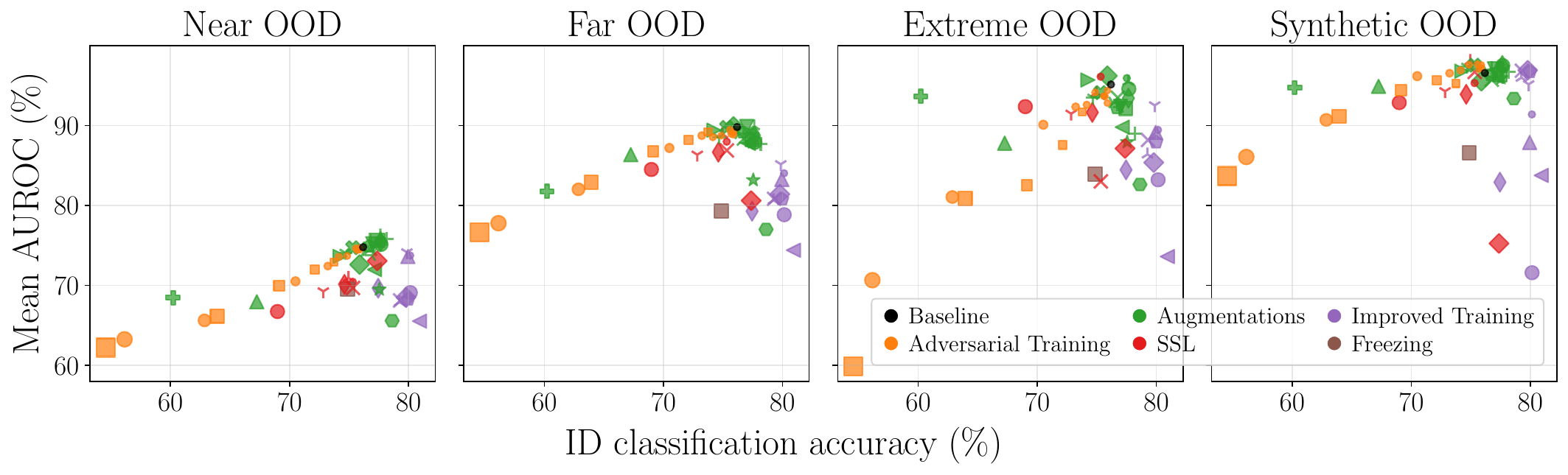}
    \caption{
    Relationship between ID accuracy and OOD detection performance (AUROC) across the OOD categories (near, far, extreme, synthetic).
    Each point represents one of the \nmodels{} ResNet-50 models, averaged over \nmethods{} OOD detection methods.
    Color indicates the model's training category, while the marker shape uniquely identifies each model within that category.
    }
    \label{fig:auroc_vs_acc_per_ood_group}
\end{figure*}

\subsection{Problem Setting}
Ideally, a deep neural network deployed in the open world should \emph{know what it does not know}: it must flag unfamiliar out-of-distribution (OOD) inputs while maintaining high accuracy on familiar in-distribution (ID) inputs.

Formally, let $\calX \subset \mathbb{R}^{C\times H\times W}$ denote the input space and $\calY = \{\text{class}_{1}, \cdots, \text{class}_{K}\}$ the label set of a $K$-way classification problem.
The network $f_{\theta}: \calX \xrightarrow{} \mathbb{R}^{K}$ is trained once---using a specific training recipe $r$ that fixes augmentations, loss, schedule, and regularizes---on an ID dataset $\calD_{\text{ID}}=\{(\bx_\text{i}, y_\text{i})\}_{\text{i}=1}^\text{N}$ drawn from the joint distribution $P_{\text{ID}}(\bx, y)$.
Everything outside the resulting support $\calX_{\text{ID}} \subset \calX$ defines the OOD space $\calX_{\text{OOD}} = \calX \setminus \calX_{\text{ID}}$.
We target the \emph{semantic} OOD regime, where the ID and OOD label spaces are disjoint: $\calY_{\text{ID}}=\calY$ and $\calY_{\text{OOD}}\cap\calY_{\text{ID}}=\varnothing$.
At deployment, the model encounters a test-time data stream drawn from the unknown mixture $P_{\text{test}}(\bx) = (1 - \pi) P_{\text{ID}}(\bx) + \pi P_{\text{OOD}}(\bx)$ with an unknown OOD prior $\pi \in [0,1]$.
A post-hoc OOD detector therefore provides a scoring function $S: \calX \xrightarrow{} \mathbb{R}$, that, given a threshold $\lambda$, implements the decision rule

\begin{equation}
\label{eqn:ood_detector}
G(\bx,\lambda) = 
    \begin{cases}
        \text{ID}                & \text{if}\: S(\bx) \geq \lambda  \\
        \text{OOD}               & \text{otherwise}
    \end{cases}.
\end{equation}
This decision function must i) accept and maintain high classification accuracy on ID samples from $P_{\text{ID}}$ and ii) reject OOD samples from $P_{\text{OOD}}$---all post-hoc, \ie, without retraining $f_{\theta}$ or relying on OOD samples during calibration.%

\subsection{Experimental Setup}
\label{sec:experimental_setup}

\paragraph{Datasets.}
Evaluating OOD detection performance typically involves selecting a single ID dataset and multiple disjoint OOD datasets that do not share semantic classes with the ID data.
Following established large-scale benchmark protocols~\cite{yang2022openood, zhang2024openood, xu2024scaling, bitterwolf2023_ninco}, we use ImageNet~\cite{deng2009_imagenet} as our ID dataset and adopt a diverse set of standard OOD datasets for evaluation.
To systematically capture different degrees of complexity, the OOD datasets are grouped into four categories:  
i) \textit{near} (SSB-Hard~\cite{vaze2022openset}, NINCO~\cite{bitterwolf2023_ninco}), 
ii) \textit{far} (iNaturalist~\cite{Horn_2018_inat}, Textures~\cite{cimpoi14dtd}, OpenImage-O~\cite{wang2022_ood_vim})---two groupings commonly used in prior work to represent increasing semantic dissimilarity to the ID dataset; 
iii) \textit{extreme} (MNIST~\cite{lecun-mnisthandwrittendigit-2010}, Fashion-MNIST~\cite{xiao2017_fmnist}), which represent visually and structurally distant domains, and 
iv) \textit{synthetic}, synthetic unit-test datasets introduced by NINCO~\cite{bitterwolf2023_ninco} to probe specific OOD weaknesses such as sensor failures.
This setup ensures compatibility with established benchmarks~\cite{zhang2024openood} while enabling a more fine-grained analysis of model behavior across a wide spectrum of data shifts.

\paragraph{Models.}
Our analysis is conducted using models based on the ResNet-50 architecture~\cite{He2015_resnet}, all of which are trained on ImageNet~\cite{deng2009_imagenet}. 
Fixing the architecture is a central design choice, as each network family exhibits unique inductive biases, activation statistics, and feature distributions, all of which can impact OOD detection performance. 
By holding the architecture constant, we isolate the effect of the training strategy alone, thereby avoiding confounding factors such as architecture and model capacity, which can independently influence both ID accuracy and OOD detection performance, thereby complicating the interpretation of their relationship.

Restricting the study to ResNet-50 provides further benefits: i) it is one of the most widely used architectures in both research and practice, serving as the de facto standard for evaluating post-hoc OOD detection methods---particularly in large-scale settings~\cite{zhang2024openood, xu2024scaling,  hendrycks2019_ood_mls, sun2022knnood, zhang2023_she}, and ii) it remains a practical choice in many real-world applications where large-scale models are infeasible due to constraints on both training and inference resources.

All models are trained on ImageNet to avoid OOD contaminations in our benchmarks.
Importantly, none of the models were optimized for OOD detection, allowing us to investigate how the training strategy alone influences OOD detection performance, independent of any OOD-specific tuning.
In total, we evaluate \nmodels{} ResNet-50 models trained using a diverse set of strategies that affect both ID classification accuracy and generalization characteristics.
These strategies are grouped into the following six categories: 
i) \mycircle{baseline} a \textit{baseline} model with the original training~\cite{He2015_resnet}, 
ii) \mycircle{advtrain} \textit{adversarial training}~\cite{salman2020_AT_IN, madry2018_AT} against PGD adversary,
iii) \mycircle{augment} different \textit{augmentation} methods ~\cite{Cubuk2019_autoaugment, lim2019_fastaugment, geirhos2018_styleaug, NEURIPS2020_randomaug, hendrycks2020_augmix, zheng2022deepaugment, pinto2022regmixup, jaini2024intriguing, erichson24_noisymix, muller2023_opticsaug, modas22_prime, hendrycks2022robustness_pixmix, zhang2018mixup, Sangdoo2019_cutmix, li2021shapetexture},
iv) \mycircle{contrastive} \textit{contrastive learning}~\cite{Caron2021_dinov1, chen21_mocov3, chen20_simclrv2, caron20_swav, khosla20_supcon} with supervised finetuning of the classification head, 
v) \mycircle{recipes} improved \textit{training recipes}~\cite{rw2019timm, wightman2021resnet, Paszke19_pytorch, Vasilis2023_TV2}, 
and vi) \mycircle{freezing} model with \textit{randomly weighted} spatial convolutional filters~\cite{gavrikov2024the}.
To distinguish individual models within each group, we use distinct marker shapes. 
The combination of marker and color yields a unique identifier for each model configuration.
This visual encoding is consistently used throughout our study, and a complete legend is available in the supplementary material.

\paragraph{Evaluation metrics.}
Following standard OOD evaluation protocols, we primarily report the Area Under the Receiver Operating Characteristic curve (AUROC) to measure the separability between ID and OOD samples. 
For comparability and completeness, we also report the False Positive Rate at a $95\%$ True Positive Rate (FPR95) in the supplementary material.
To gain a deeper understanding of model behavior, we also follow the diagnostic metrics used in~\cite{Humblot-Renaux_2024_CVPR}: 
$\text{AUROC}_{\text{correct~vs.~OOD}}$ and $\text{AUROC}_{\text{incorrect~vs.~OOD}}$. 
These metrics isolate the OOD detection performance on correctly and incorrectly classified ID samples, respectively. 
Ideally, a robust OOD detector should perform well on both, signaling that its scores are not merely correlated with classification correctness but reflect true distributional shifts.
Finally, we also report the $\text{AUROC}_{\text{correct~vs.~incorrect}}$, which measures a method’s ability to flag misclassified ID samples~\cite{bernhardt2022_failuredetection, Humblot-Renaux_2024_CVPR}.

\begin{figure*}
    \centering
    \includegraphics[width=\linewidth]{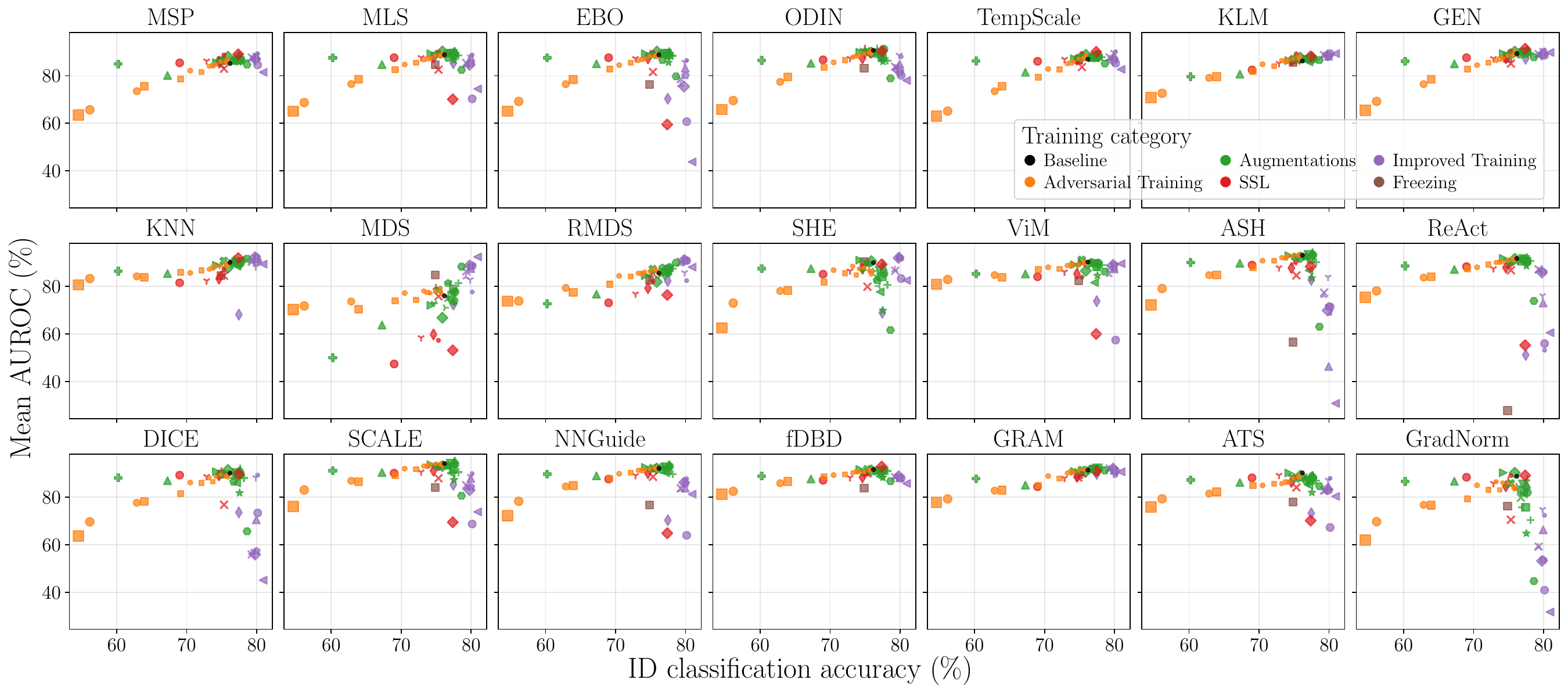}
    \caption{
    Relationship between ID accuracy and OOD detection performance (AUROC) for each of the \nmethods OOD detection methods.
    Each point represents one of the \nmodels{} ResNet-50 models, averaged over \noodsets{} OOD datasets.
    Color indicates the model's training category, while the marker shape uniquely identifies each model within that category. 
    Best viewed on screen.
    }
    \label{fig:auroc_vs_acc_per_ood_method}
\end{figure*}

\paragraph{Test time and evaluation.}
At test time, the inputs are resized to $256\times256$ and center-cropped to $224\times224$.
We evaluate \nmethods{} approaches using the OpenOOD benchmark~\cite{yang2022openood, zhang2023openood, zhang2024openood}, a comprehensive and standardized open-source framework for benchmarking state-of-the-art OOD detection methods.
To facilitate analysis, we categorize the OOD detection methods into five groups based on the type of information they leverage to compute the OOD score:
i) \mycircle{classification} \emph{classification-based} methods primarily use model outputs such as softmax probabilities or logits (MSP~\cite{hendrycks2016_ood_msp}, MLS~\cite{hendrycks2019_ood_mls}, EBO~\cite{liu2020_ood_ebo}, ODIN~\cite{shiyu17}, TempScale~\cite{Gua2017_tempscaling}, KLM~\cite{hendrycks2019_ood_mls}, GEN~\cite{Liu2023_GEN});
ii) \mycircle{feature} \emph{feature-based} methods derive the score from the penultimate layer (MDS~\cite{lee2018_mahala}, RMDS~\cite{Ren2021_rmds}, KNN~\cite{sun2022knnood}, SHE~\cite{zhang2023_she})
iii) \mycircle{hybrid} \emph{hybrid} methods combine information from both the output and penultimate layer (ViM~\cite{wang2022_ood_vim}, ASH~\cite{djurisic2023ash}, ReAct~\cite{sun2021_ood_react}, DICE~\cite{sun2022dice}, SCALE~\cite{xu2024scaling}, NNGuide~\cite{park2023nnguide}, fDBD~\cite{liu2024fast}),
iv) \mycircle{inter} \emph{intermediate-feature} methods use information from shallow to deep layers (GRAM~\cite{sastry20a_gram}, ATS~\cite{Krumpl2024_ats}), and
v) \mycircle{gradients} \emph{gradients}  methods utilize gradients (GradNorm~\cite{huang2021_gradnorm}).
For methods that require ID data to compute statistics or calibration parameters, we use the training split of the ID dataset. 
For methods involving hyperparameter tuning, we rely on a held-out validation set comprising both ID and OOD samples to ensure fair and robust evaluation.
Further implementation details, including tuning procedures and parameter settings, are provided in the supplementary material.

\section{Analysis}
\label{sec:analysis}
Building on our large-scale experimental setup, we characterize the key determinants of OOD detection performance by addressing four key questions:
i) Does higher ID accuracy imply better OOD detection? ii) Are OOD detectors merely identifying misclassified samples? iii) Where does AUROC variance originate from? and iv) How robust are detection methods across training variants?

\paragraph{Does higher ID accuracy imply better OOD detection?}
Our analysis reveals that the relationship between ID accuracy and OOD detection performance, averaged across all OOD detection methods, is not monotonic but rather a distinct \emph{rise-then-fall} pattern (\cref{fig:teaser}), challenging the common assumption of a simple positive correlation. 
Initially, as accuracy improves from lower levels up to the vanilla ResNet-50 baseline ($76.19\%$), a positive correlation exists (Spearman's $\rho=0.38$, $p \ll 0.001$). 
This region is primarily populated by models subjected to adversarial training (orange cluster), which degrades ID accuracy while compressing logit margins.
This trend aligns with the findings of prior work~\cite{Humblot-Renaux_2024_CVPR}, which observed similar behavior when analyzing the impact of label noise, where models also fell into a low-to-baseline accuracy region.

However, once accuracy reaches or exceeds the baseline band, the relationship reverses and becomes weakly negative---AUROC declines slightly as accuracy rises (Spearman's $\rho=-0.082$, $p \ll 0.001$).
A global correlation over all models and methods fails to reveal this non-monotony, yielding a coefficient near zero (Spearman's $\rho=0.04,\ p \ll 0.001$).
This rise-then-fall pattern is consistent across all OOD dataset categories (\cref{fig:auroc_vs_acc_per_ood_group}), with near-OOD showing a smaller min--max spread, while extreme-/synthetic-OOD exhibit larger spreads across methods and models (Supp.~Tab.~4).

At the OOD detection method level, the picture is even more diverse (\cref{fig:auroc_vs_acc_per_ood_method}): the \emph{rise} phase is steep for confidence-based scores such as MSP and GEN, and modest for distance metrics like KNN, and fDBD, while the \emph{fall} ranges from pronounced (SCALE, ReAct) to negligible (GEN, GRAM) or even non-existent (KLM).
Such heterogeneity demonstrates that OOD performance depends strongly on the interaction between the OOD detection method heuristic and the training strategy, highlighting that the relationship between OOD detection and ID accuracy is complex.

The downturn at higher accuracies indicates that training strategies designed to maximize ID accuracy and generalization can undermine post-hoc OOD detection; as shown in the supplementary, aggressive regularization (MixUp, CutMix, TorchVision 2) compresses and sparsifies the feature space and narrows max-logit distributions, increasing ID-OOD overlap.
Consequently, methods that rely on model output or specific activation characteristics degrade more, whereas geometry/statistics-based methods remain comparatively stable (see Supp.~App.~B and Supp.~Fig.~18--20).%

We contend that these complex effects were previously overlooked because prior studies, which often did not include models trained with diverse strategies, focused on a limited subset of OOD methods or had results confounded by variations in model architecture and capacity.

\begin{figure}[t]
  \centering
   \includegraphics[width=1.0\linewidth]{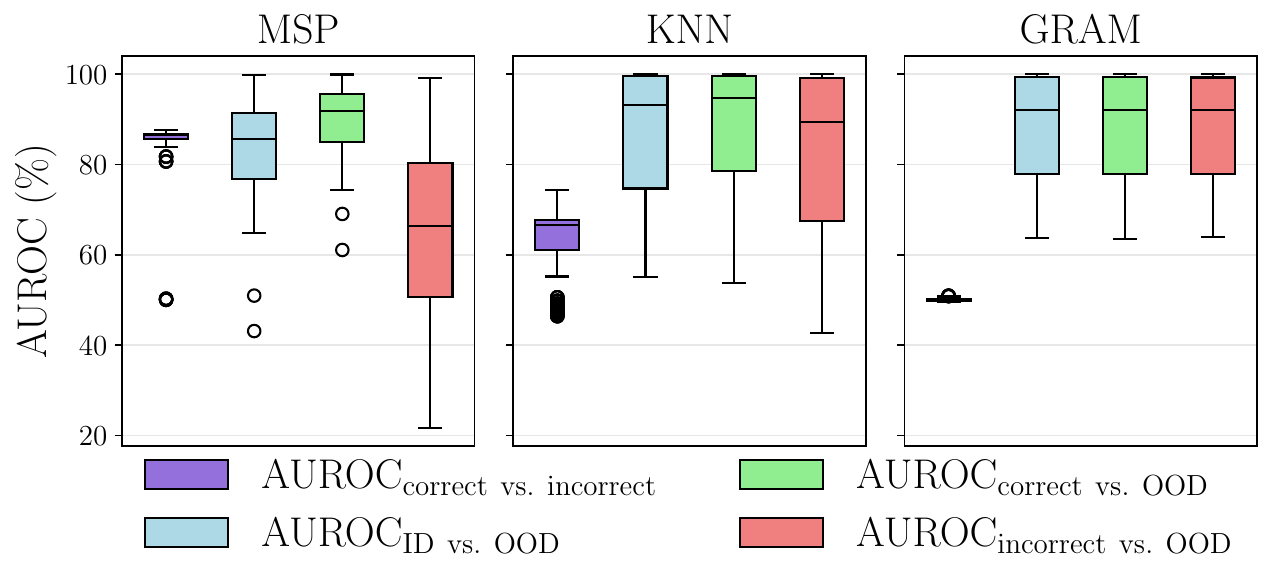}

   \caption{
        Comparing the performance of MSP, KNN, and GRAM across multiple AUROC-based evaluation metrics.
        $\text{AUROC}_{\text{correct~vs.~incorrect}}$ evaluates failure prediction on ID data only, distinguishing between correctly and incorrectly classified samples. 
        The remaining metrics assess OOD detection, either across all ID samples, only correctly classified ones, or only misclassified ones. 
        Each boxplot shows the distribution over \nmodels{} models and \noodcategories{} OOD categories.
   }
   \label{fig:main_paper_methods_auroc_boxplots}
\end{figure}

\paragraph{Are OOD detectors merely identifying misclassified samples?}

Prior work~\cite{Humblot-Renaux_2024_CVPR} suggested that the correlation between ID accuracy and OOD performance might arise because detectors simply distinguish correctly classified ID inputs from OOD samples, while misclassifications are often confused with OOD. 
We test this hypothesis by analyzing OOD detector performance separately on correctly and incorrectly classified ID subsets.

\begin{figure}[t]
  \centering
   \includegraphics[width=1.0\linewidth]{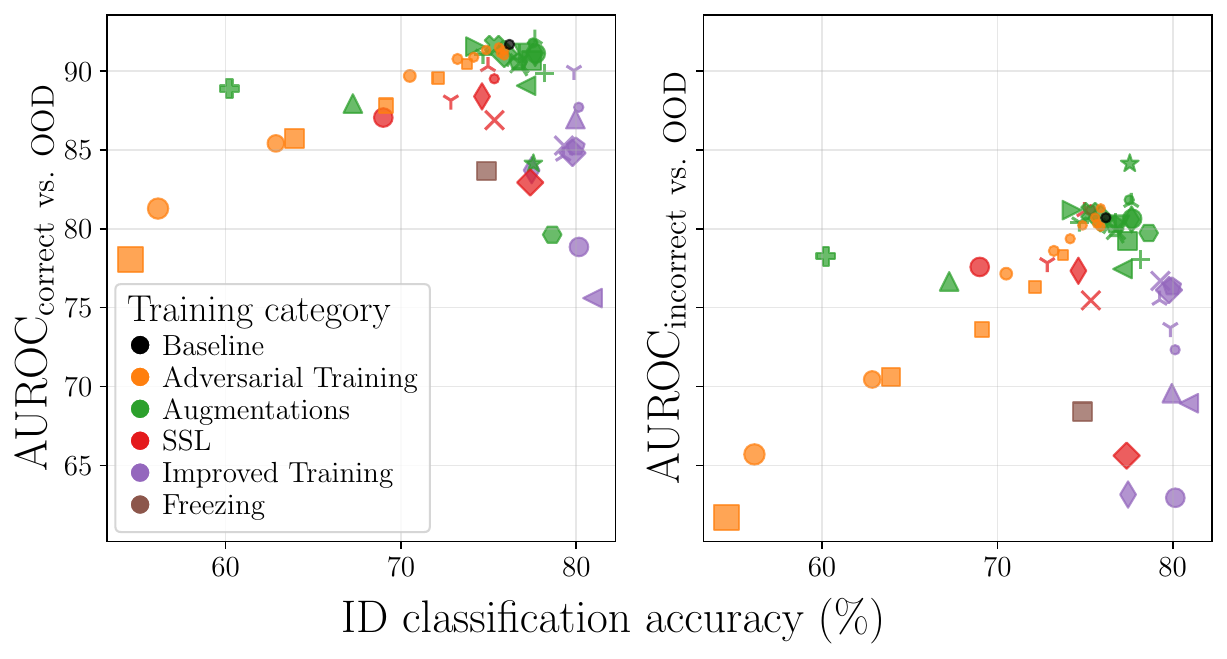}

   \caption{
    Relationship between ID classification accuracy and OOD detection performance.
    We distinguish between the ability of OOD detectors to separate correctly classified ID samples from OOD samples (left), and incorrectly classified ID samples from OOD samples (right).
    Each point represents one of the \nmodels{} ResNet-50 models, averaged over \nmethods{} OOD detection methods and \noodcategories{} OOD categories.
    Color show the model's training category; marker shapes uniquely identify models within each category.
   }
   \label{fig:correct_incorrect_ood_auroc_vs_accurracy}
\end{figure}

\cref{fig:correct_incorrect_ood_auroc_vs_accurracy} shows that $\text{AUROC}_{\text{correct vs.~OOD}}$ and $\text{AUROC}_{\text{incorrect vs.~OOD}}$ follow the same \emph{rise-then-fall} pattern observed in our global analysis (\cref{fig:teaser}).
The two metrics are strongly correlated (Spearman's $\rho = 0.87,\ p \ll 0.001$), indicating that methods effective at distinguishing correctly classified ID samples from OOD data generally remain robust even when the ID sample is misclassified.
A persistent gap of roughly $10\%$ separates the two scores, with the largest discrepancy for confidence-based methods that compute their OOD scores directly from model output. 

\cref{fig:main_paper_methods_auroc_boxplots} contrasts three representative methods: MSP, KNN, and GRAM.
MSP is highly sensitive to correctness: its failure-detection score is high ($\text{AUROC}_{\text{correct vs.~incorrect}} = 84.75\%$), and its OOD AUROC falls sharply from $\text{AUROC}_{\text{correct vs.~OOD}} = 90.72\%$ on correctly predicted ID samples to $\text{AUROC}_{\text{incorrect vs.~OOD}} = 67.70\%$ on misclassified ones.
KNN shows a lower failure detection AUROC ($63.15\%$) with a much narrower gap (correct: $89.13\%$, incorrect: $83.96\%$), resulting in a higher overall AUROC of $87.93\%$.
GRAM achieves virtually identical scores on both subsets (correct: $89.67\%$, incorrect: $89.66\%$), with a near-random failure-detection AUROC ($50.03\%$), demonstrating robustness to ID classification quality.

The supplementary material confirms this pattern across all detectors: methods that derive their scores from richer feature representations (\eg, KNN, NNGuide, and GRAM) are less sensitive to misclassified ID samples, whereas confidence-based approaches that rely solely on the model output experience the largest performance drop, with their sensitivity varying substantially across training recipes.

\paragraph{Where does AUROC variance originate from?}

~\cref{fig:teaser,fig:auroc_vs_acc_per_ood_method} reveals large performance variations across OOD detectors and training strategies, prompting a formal variance analysis.
To quantify factors contributing to AUROC variability, we run a three-way ANOVA with factors model (\ie, training strategy), method (\ie, OOD detection method), and OOD dataset category (\ie, OOD category).

\begin{figure}[t]
  \centering
   \includegraphics[width=1.0\linewidth]{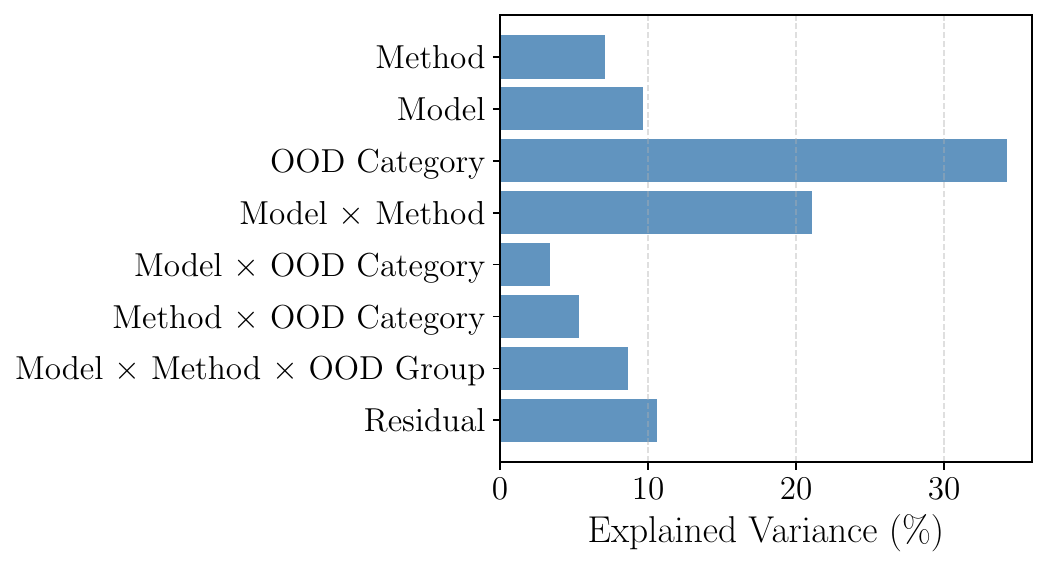}

   \caption{
        Proportions of AUROC variance explained by each factor in the three-way ANOVA (\nmodels models $\times$ \nmethods OOD detection methods $\times$ 4 OOD categories).
   }
   \label{fig:anova_effects}
\end{figure}

\cref{fig:anova_effects} shows that the OOD category explains the largest share of variance ($34.22\%$), reflecting the intrinsic difficulty gap between the OOD categories (near, far, extreme, and synthetic).
The model--method interaction ranks second ($21.05\%$), far exceeding the main effects of either factor alone; detector performance thus depends strongly on how its heuristic aligns with the feature statistics induced by a given training recipe.
A smaller yet non-trivial three-way interaction ($8.66\%$) and a residual term ($10.62\%$) indicate systematic rather than random effects. 

Robustness checks confirm this picture.
Omitting one OOD category at a time (details in the supplementary material) shows that removing the toughest category (near-OOD) cuts the OOD category term and boosts the
model--method interaction to $33.4\%$.
The coupling between detector and model is therefore partially masked when all methods struggle uniformly on the most challenging OOD test data.
Across all runs, the interaction term remains a major contributor, underscoring its central importance.

In summary, OOD performance depends as much on the alignment between the OOD detector and the model as on the OOD set itself, underscoring the critical role of the training strategy.

\paragraph{How robust are detection methods across training variants?}

A central finding of this work is that the performance of an OOD detector is conditional: it depends not only on the OOD data but also on the training strategy used to obtain the model.
OOD detection methods are typically validated on a single training strategy, often the ResNet-50 baseline~\cite{He2015_resnet}.
This means that there is no universally best method, but it also implies that selecting the most suitable method for a given task can be a challenge.
Robustness, defined here as delivering consistently strong OOD detection performance across heterogeneous training strategies, therefore becomes as important as
raw performance.

\begin{figure}[t]
  \centering
   \includegraphics[width=1.0\linewidth]{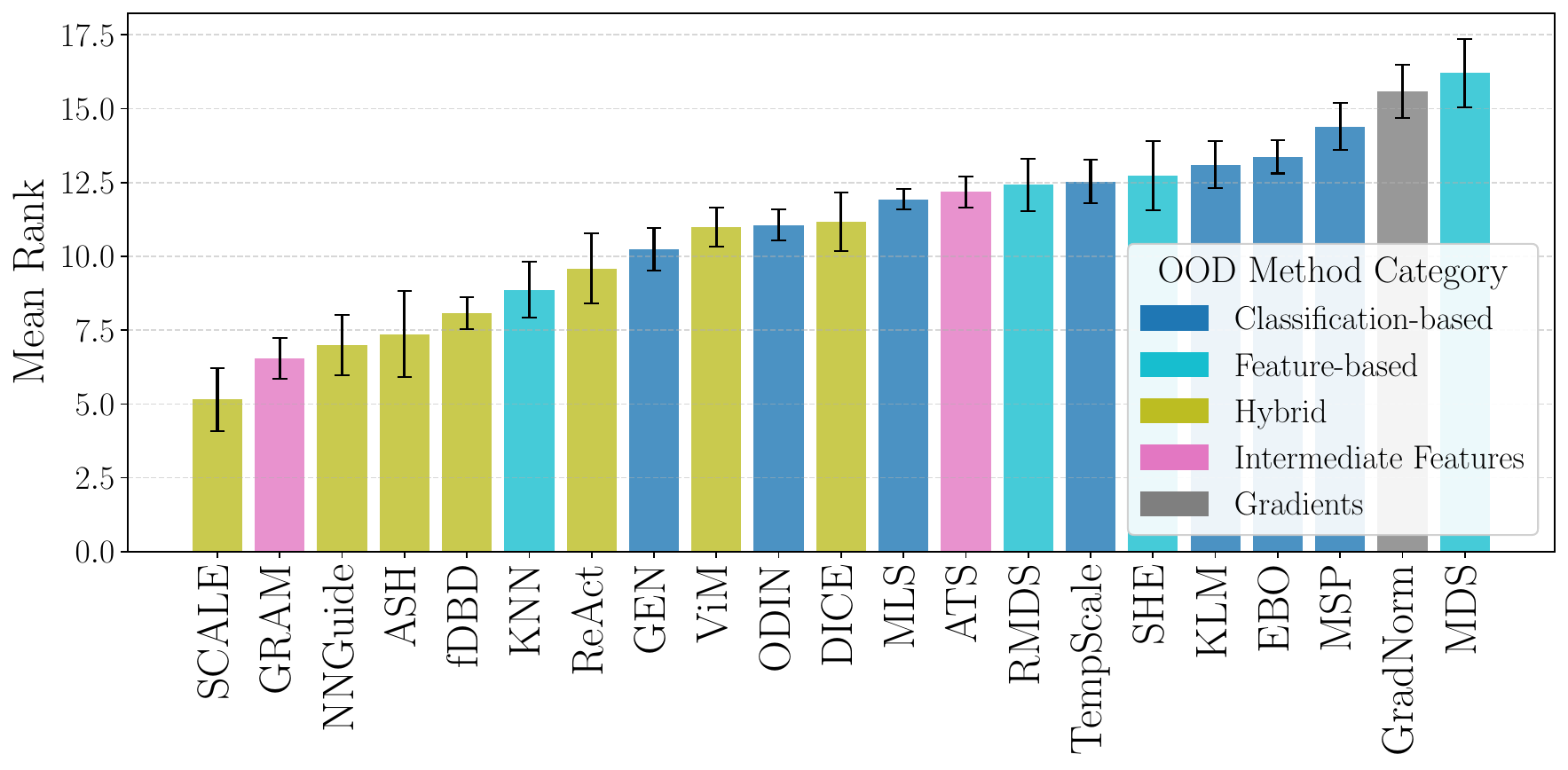}

   \caption{
        Robustness of OOD detection methods measured by model-averaged mean rank across \nmodels{} models. %
        Methods are ranked per model by mean AUROC, and the overall rank is the average across models.
        Error bars show $95\%$ CIs; lower rank is better.
   }
   \label{fig:robust_ranking}
\end{figure}

To consider both aspects, \cref{fig:robust_ranking} orders all OOD detection methods by their mean rank over the \nmodels models.
A Friedman test confirms statistically significant differences among methods ($\chi^2 = 467.97$, $p \ll 0.001$), validating this analysis.
Model enhancement OOD detection methods dominate: SCALE leads (mean rank $5.15$, $95\%$ CI $[4.08,\ 6.22]$), followed by NNGuide, ASH, and fDBD.
The intermediate-feature-based method GRAM, introduced in 2020, also proves to be a strong and consistent performer, ranking second with a tight confidence interval (mean rank $6.54$, 95\% CI $[5.84,\ 7.24]$).
In contrast, several classification-based methods, such as MSP, KLM, and TempScale, occupy lower ranks on average.

\begin{figure}[t]
  \centering
   \includegraphics[width=1.0\linewidth]{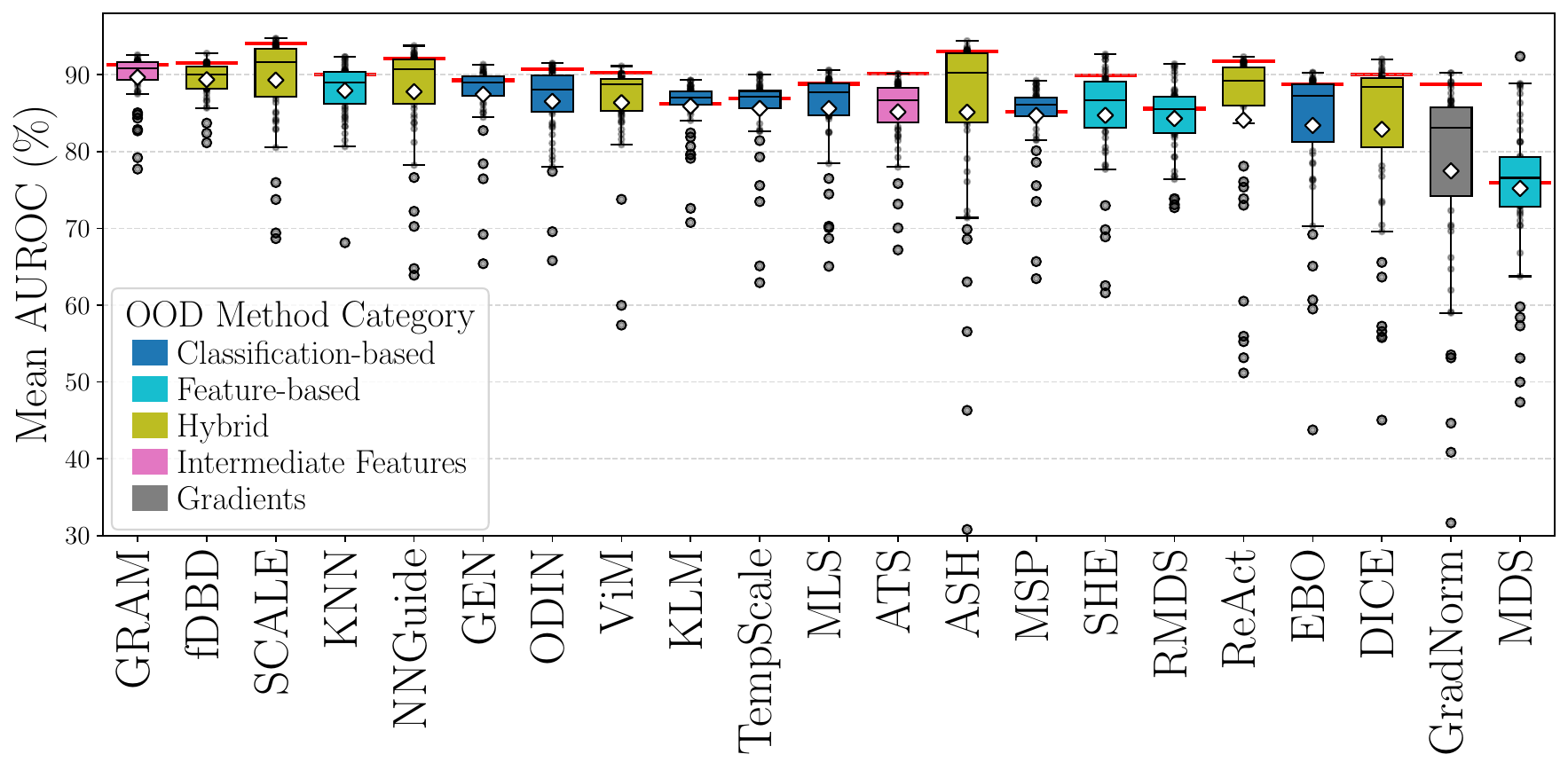}

   \caption{
       Mean AUROC distribution for each OOD detection method, across \nmodels{} models and \noodcategories{} OOD categories.
       Diamond markers denote the mean for each method; the red horizontal line indicates that method's performance on the baseline ResNet-50~\cite{He2015_resnet}.
       Methods are sorted by overall mean AUROC.
   }
   \label{fig:auroc_dist_per_method}
\end{figure}

\Cref{fig:auroc_dist_per_method,fig:auroc_vs_acc_per_ood_method} reveal two further insights.
First, many of the top-ranked methods exhibit high variance, meaning that while their performance can be excellent, it can also be mediocre when paired with the wrong training strategy. 
Second, many models, such as ASH, SCALE, and ReAct, as well as some of the weaker performers, including DICE and GradNorm, exhibit their best or nearly optimal performance with the baseline model.
This pattern suggests an over-specialization to the characteristics of vanilla training, on which they were likely developed and benchmarked.

These two observations highlight a trade-off between high-performing specialists and more consistent generalists. 
The specialists, such as ReAct and SCALE, are effective because their heuristics are finely tuned to specific characteristics of vanilla-trained models, for instance, clipping high-activation features. 
This explains both their SOTA performance on the baseline and their high variance across diverse training strategies, as advanced training recipes alter the very activation patterns on which these methods depend. 
Generalists leverage more invariant properties.  
GRAM and fDBD illustrate this: they remain among the top methods (\cref{fig:robust_ranking}), show only modest performance drops from baseline (GRAM: $-1.74\%(\pm3.24)$, fDBD: $-2.17\% (\pm 2.38)$), and maintain tight score distributions (\cref{fig:auroc_dist_per_method}). 
GRAM’s higher-order statistics and fDBD’s distance metric depend less on absolute activation magnitudes, making them less sensitive to training-induced changes.

OOD robustness hinges primarily on one key factor: how well a detector’s scoring rule aligns with the feature representations produced by a given training recipe. 
Stability is achieved not by richer features alone, but by combining them with training-agnostic heuristics (\eg, statistical summaries or distance-based measures) that remain effective across the diverse characteristics and activation patterns induced by varying training strategies.

\section{Discussion and Conclusion}

Our study challenges, extends, and aligns with previous findings on the relationship between ID classification accuracy and OOD detection performance. 
Contrary to simplified claims of a universally positive or negative correlation, our large-scale, controlled analysis demonstrates a non-monotonic and method-dependent relationship for OOD detection.
While prior work has often reported a positive correlation between ID accuracy and OOD detection performance, these findings frequently stem from studies with significant potential confounders. 
For instance, studies \cite{vaze2022openset, galil2023fwbm} evaluated a broad range of architectures combined with only a small subset of OOD detection methods, while \cite{Humblot-Renaux_2024_CVPR} varied both datasets and architectures.
In contrast, a recent study~\cite{wang2024dissect} reported negative correlations for large-scale benchmarks, examining narrowly optimized model sets in the context of outlier exposure and thus overlooking the initial positive trend at lower accuracy ranges. 
Our results clarify these apparent contradictions by revealing them as different perspectives on the same nuanced relationship.

Critically, we reveal a strong interaction effect between models and OOD detection methods, which alone accounts for over $20\%$ of OOD detection performance variance. 
This suggests that the effectiveness of an OOD detection method is not an independent property, but rather is fundamentally coupled with the specific, training-induced characteristics of the model. 
This high degree of interaction has direct and critical implications for our field. 
Evaluating OOD methods on a single model configuration captures only a limited slice of a more extensive interaction space, potentially rewarding overly specialized methods. 
Our findings issue a clear call for a shift in evaluation culture.
Moving forward, robust benchmarking must adopt multi-model evaluations across a wide spectrum of training strategies to establish a method's true generalization, reveal its limitations and hidden dependencies, and ensure the community is not simply optimizing for methods that are brittle and over-specialized to a single, default training condition.

In summary, our findings demonstrate that a simple monotonic correlation does not hold, confirming that ID accuracy alone is not a reliable indicator for OOD detection performance. 
Thus, the assumption that improved ID accuracy automatically translates into better (or worse) OOD detection performance is not justified.
Stability and robust OOD generalization result instead from a method’s ability to combine rich, training-induced features with patterns that remain stable across diverse training strategies.

\paragraph{Guidance for Practitioners.}
Practitioners should be aware that while achieving baseline levels of ID accuracy generally supports better OOD detection, further accuracy improvements through sophisticated training methods do not universally guarantee enhanced OOD detection performance, and can instead harm it.
When utilizing standard, vanilla models, top-ranked methods from established benchmarks typically remain reliable. 
However, for models customized with advanced training or strong regularization, generalist OOD detectors should be preferred that do not rely on specific activation patterns---in practice, geometry/statistics-based methods that use distances or higher-order feature statistics (\eg, KNN, GRAM, RMDS) are typically more stable; otherwise, tailored evaluations should be performed to select the OOD detector best aligned with the model's learned representations.

\paragraph{Limitations and Future Work.}

A central limitation of our study is the deliberate restriction to the ResNet-50 architecture. 
This methodological decision was necessary to isolate training-induced effects from architectural or capacity-based confounding factors, a challenge in prior studies.
While this design provides clear and definitive insights, our findings also support recent hypotheses~\cite{zhang2024openood} that OOD methods are implicitly tuned to CNN-based architectures, explaining why alternatives such as Vision Transformers (ViTs)~\cite{dosovitskiy2021vit}, despite higher ID accuracy, often show weaker OOD performance.
While a preliminary ablation study (App.~C) confirms our main findings also for ViTs, a similarly detailed examination of the model-method interactions is a valuable future direction.

\paragraph{Conclusion.}
\label{sec:conclusion}

We have identified important shortcomings in previous studies that show correlations between ID accuracy and OOD performance.
Our large-scale, carefully designed study shows that while correlations exist between certain ID metrics and OOD performance under specific conditions, OOD robustness ultimately depends on the interaction between the model and the detection method, and these correlations do not reflect the broader dynamics of generalization.
This makes OOD performance too complex to be inferred from a single ID measure. 
This calls for a shift toward systematic evaluation across diverse models and the development of diagnostics that capture model--method compatibility---essential for reliable OOD detection and trustworthy pre-deployment assessment.

\textbf{Acknowledgments}
This work was supported by KESTRELEYE GmbH, whose financial contribution and commitment made this research possible.

{
    \small
    \bibliographystyle{ieeenat_fullname}
    \bibliography{main}
}

\appendix
\clearpage
\maketitlesupplementary

\crefname{appendix}{Appendix}{Appendices}
\Crefname{appendix}{Appendix}{Appendices}
\crefalias{section}{appendix}

This supplementary document expands on the main manuscript. 
It provides full experimental details (\cref{sec:supp:experimental_details}), comprehensive results that support and extend our analyses (\cref{sec:supp:supp:detailed_results}), and additional experiments with Vision Transformers (ViT) (\cref{sec:supp:result_vit}).

\section{Experimental Details}
\label{sec:supp:experimental_details}

\subsection{Implementation Details}
\paragraph{Software stack.}
Our experimental framework is built upon the OpenOOD~\cite{yang2022openood, zhang2023openood, zhang2024openood} framework. 
Specifically, we utilize the public fork from Humblot-Renaux~\cite{Humblot-Renaux_2024_CVPR}, as it provides a GRAM~\cite{sastry20a_gram} implementation that follows the official implementation details, which also leverages information from intermediate layers. 
We extend the model zoo by integrating \nmodels ImageNet checkpoints, adapted from the collection provided by~\cite{Gavrikov2024_biases_generalization}.
The full list of all models used in this study, along with their training categories and performance metrics, is detailed in \cref{tab:model_list}.
All evaluations in this study are executed within this unified framework.

To broaden the evaluation scope, we also enrich the data layer with two additional OOD categories:
(i) \emph{extreme-OOD} including MNIST~\cite{lecun-mnisthandwrittendigit-2010} and Fashion-MNIST~\cite{xiao2017_fmnist}, and
(ii) \emph{synthetic-OOD} including the unit-test data provided by NINCO~\cite{bitterwolf2020_card}.
This setup ensures reproducibility and fair comparison with a broad set of diverse training strategies, OOD test sets, and existing state-of-the-art OOD detection methods. 

\paragraph{Hardware and system configuration.}
All experiments were executed on a workstation equipped with an Intel Core i9-9900X (10 cores, 3.5 GHz) and two NVIDIA GPUs (RTX 2080Ti \(+\) RTX 3090).
The software environment consisted of Ubuntu 22.04, Python 3.10, PyTorch 2.0.1, and CUDA 11.8.

\paragraph{OOD detection methods.}
Many OOD detection methods require a configuration phase prior to evaluation, for which we strictly follow the OpenOOD benchmark protocols to ensure comparability. 
This process includes two types of setup: some methods are calibrated on the ID training set to compute statistics or other parameters, while others have crucial hyperparameters that are tuned on a held-out validation set containing both ID and OOD samples. 
Although detectors ship with default hyperparameters, these defaults are typically tuned to a vanilla training recipe, which can risk biasing the comparison.  
Re-optimizing all parameters, therefore, provides a fair test across the diverse training strategies evaluated here. 
The exact settings and hyperparameter search spaces adopted for each method are detailed in ~\cref{tab:ood_hyperparameter}.

\begin{table}
\centering
\footnotesize
\resizebox{\linewidth}{!}{
\begin{tabular}{ll}
\toprule
\textbf{Method} & \multicolumn{1}{c}{\textbf{Hyperparameter Search Space}} \\
\midrule
\rowcolor{gray!10} MSP~\cite{hendrycks2016_ood_msp}             &  \\
MLS~\cite{hendrycks2019_ood_mls}                                & \\
\rowcolor{gray!10} EBO~\cite{liu2020_ood_ebo}                   & \\
 & temperature $T \in \{1, 10, 100, 1000\}$       \\
\multirow{-2}{*}{ODIN~\cite{shiyu17}} & perturbation mag. $\sigma \in \{0.0014, 0.0028\}$ \\                          
\rowcolor{gray!10} TempScale~\cite{Gua2017_tempscaling}         &  \\
KLM~\cite{hendrycks2019_ood_mls}                                &  \\
\rowcolor{gray!10}
    & gamma $\in \{0.01, 0.1, 0.5, 1, 2 ,5 ,10\}$   \\
  \multirow{-2}{*}{\cellcolor{gray!10}GEN~\cite{Liu2023_GEN}}   & \cellcolor{gray!10} top-M classes $\in \{10, 50, 100, 200, 500, 1000\}$ \\
KNN~\cite{sun2022knnood}                                        & $K \in \{50, 100, 200, 500, 1000\}$ \\
\rowcolor{gray!10} MDS~\cite{lee2018_mahala}                    &  \\
RMDS~\cite{Ren2021_rmds}                                        &  \\
\rowcolor{gray!10} SHE~\cite{zhang2023_she}                     &  \\
ViM~\cite{wang2022_ood_vim}                                     & \\
\rowcolor{gray!10} ASH~\cite{djurisic2023ash}                   & percentile $\in \{65, 70, 75, 80, 85, 90, 95\}$  \\
ReAct~\cite{sun2021_ood_react}                                  & percentile $\in \{70, 80, 85, 90, 95, 99\}$ \\
\rowcolor{gray!10} DICE~\cite{sun2022dice}                      & percentile $\in \{10, 30, 50, 70, 90\}$ \\
SCALE~\cite{xu2024scaling}                                      & percentile $\in \{65, 70, 75, 80, 85, 90, 95\}$ \\
\rowcolor{gray!10} NNGuide~\cite{park2023nnguide}               & \\
fDBD~\cite{liu2024fast}                                         & normalization $\in \{ \text{true}, \text{false} \}$ \\
\rowcolor{gray!10} GRAM~\cite{sastry20a_gram}                   & \\
ATS~\cite{Krumpl2024_ats}                                       & \\
\rowcolor{gray!10} GradNorm~\cite{huang2021_gradnorm}                              & \\
\bottomrule
\end{tabular}
}
\caption{
Overview of hyperparameter search space for all considered OOD detection methods.
}
\label{tab:ood_hyperparameter}
\end{table}

\input{tables/model_overview}

\section{Detailed Results}
\label{sec:supp:supp:detailed_results}

\begin{figure}[t]
  \centering
   \includegraphics[width=1.0\linewidth]{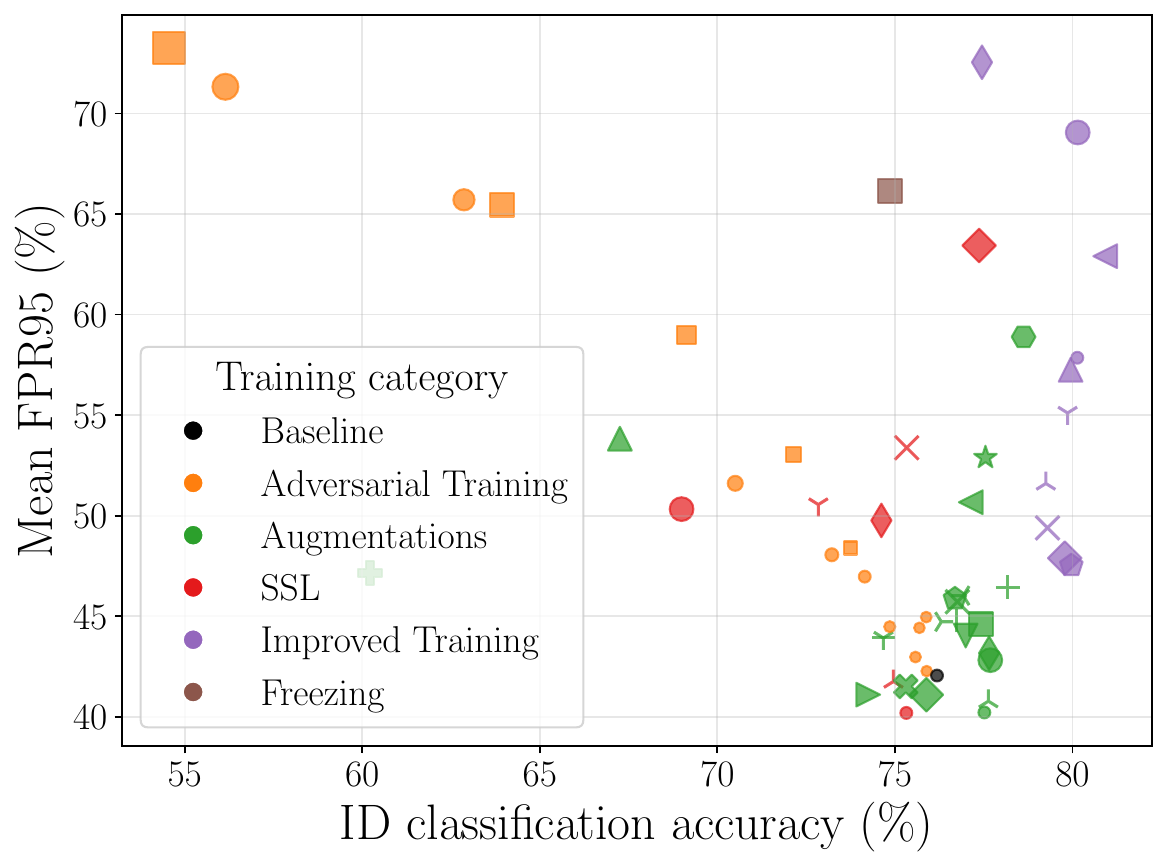}

   \caption{
    Relationship between ID classification accuracy and OOD detection performance, measured by the mean False Positive Rate at 95\% True Positive Rate (FPR95). 
    Each point represents one of \nmodels ResNet-50 models trained with a diverse strategy. 
    The reported FPR95 for each model is the average across all \nmethods OOD detection methods and \noodcategories OOD categories. 
    Color indicates the model's training category, while the marker shape uniquely identifies each model within that category.
   }
   \label{fig:fpr95_vs_acc}
\end{figure}

\paragraph{Does higher ID accuracy imply better OOD detection?}
To validate that our findings are not an artifact of the AUROC metric, we perform an equivalent analysis using the False Positive Rate at $95\%$ True Positive Rate (FPR95). 
As shown in \cref{fig:fpr95_vs_acc}, this analysis plots ID classification accuracy against FPR95, where lower values signify better OOD detection performance.

This analysis quantitatively confirms the visually observed mirrored fall-then-rise pattern. 
Consistent with the AUROC results, the overall relationship between accuracy and FPR95 yields a weak global correlation (Spearman's $\rho = -0.04, p \ll 0.001$).
Similar to the AUROC analysis, in the low-to-baseline accuracy regime, performance is primarily driven by adversarially trained models, which exhibit a strong negative correlation between ID classification accuracy and FPR95 (OOD performance improves). 
Conversely, for high-performing models, advanced augmentations and regularization techniques reverse this relationship, leading to a degradation in OOD 
performance (an increase in FPR95).

This result provides strong evidence that the complex, non-monotonic relationship between ID accuracy and OOD performance is a general phenomenon, independent of the evaluation metric.

\paragraph{}

\begin{figure}[t]
  \centering
   \includegraphics[width=1.0\linewidth]{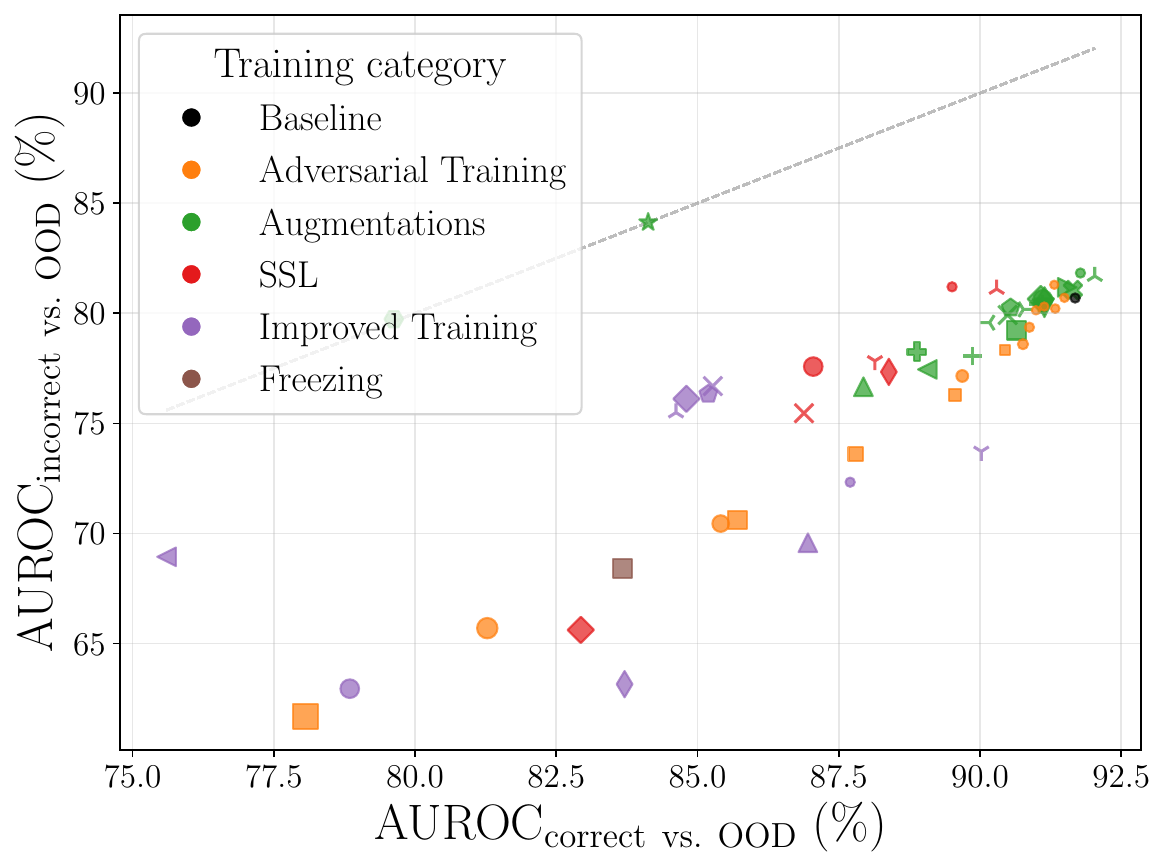}

    \caption{
        Relationship between $\text{AUROC}_{\text{correct~vs.~OOD}}$ and  $\text{AUROC}_{\text{incorrect~vs.~OOD}}$.
        Each point represents one of \nmodels models, with performance averaged across all \nmethods OOD detection methods and \noodcategories OOD categories. 
        Color indicates the model's training category, while the marker shape uniquely identifies each model within that category.
    }
   \label{fig:incorrect_correct_auroc}
\end{figure}

\begin{figure*}
    \centering
    \includegraphics[width=\linewidth]{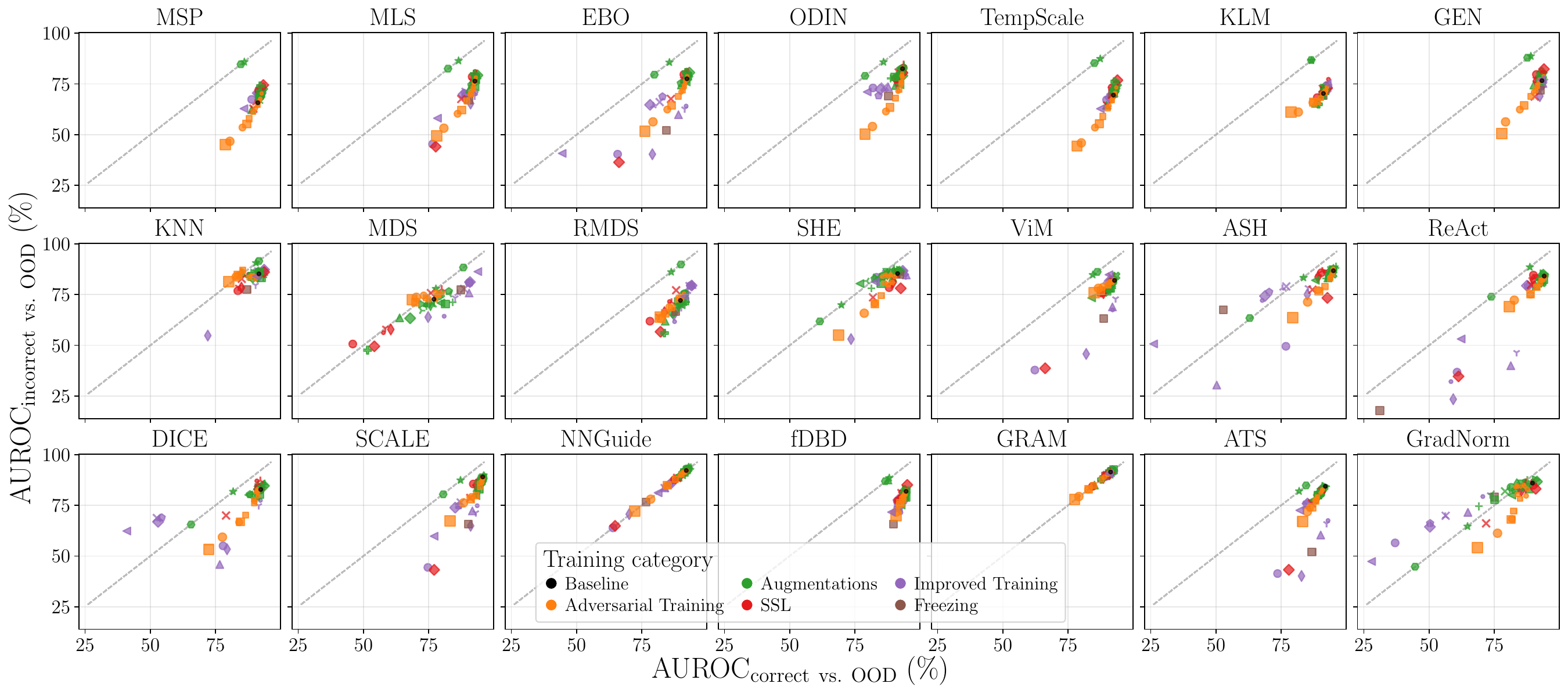}
    \caption{
    Relation between OOD Detection Performance on correct versus incorrect ID samples for each OOD detection method. 
    Each point represents one of the \nmodels{} ResNet-50 models, averaged over \noodsets{} OOD datasets.
    Color indicates the model's training category, while the marker shape uniquely identifies each model within that category.
    }
    \label{fig:correct_vs_incorrect_auroc_per_method}
\end{figure*}

\begin{figure*}
    \centering
    \includegraphics[width=\linewidth]{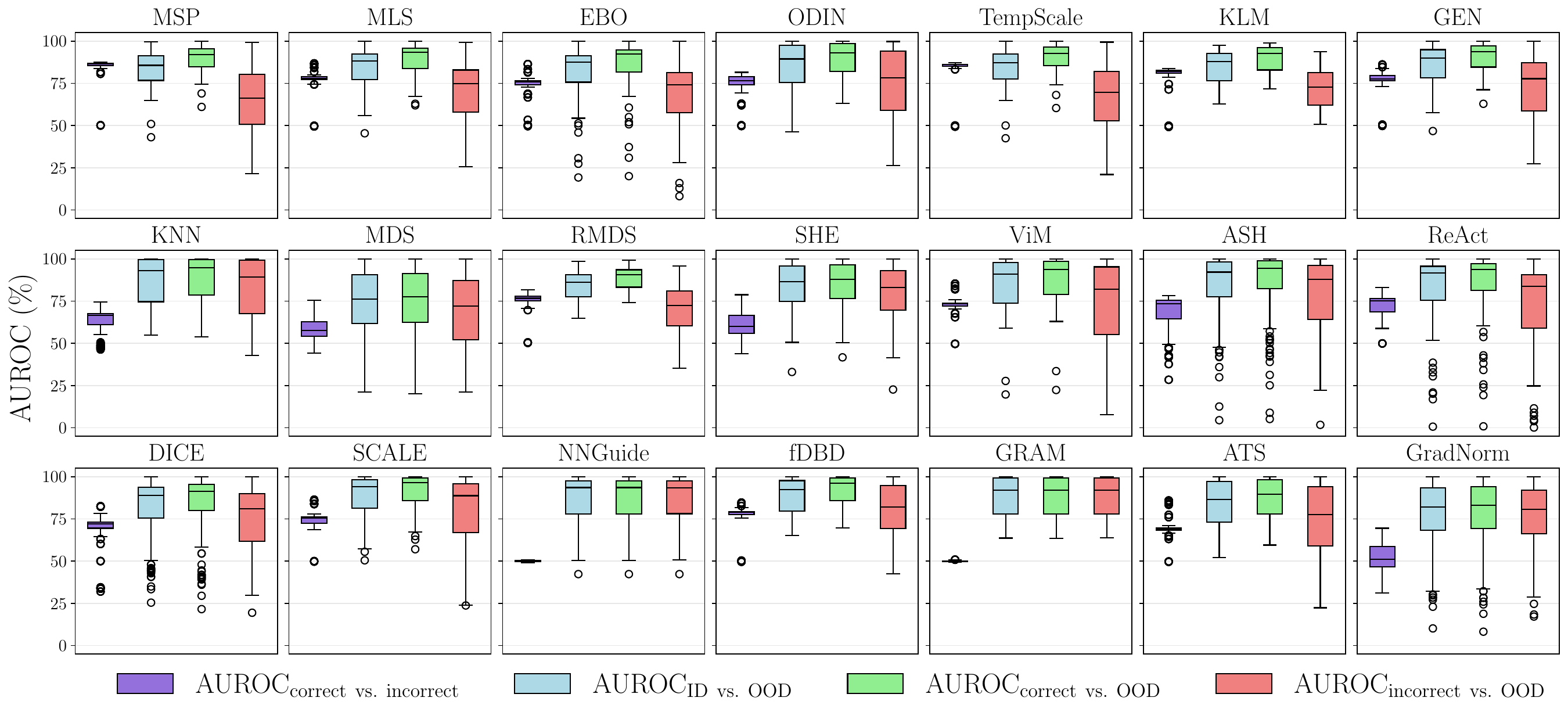}
    \caption{
    Performance comparison of all \nmethods OOD detection methods across multiple AUROC-based evaluation metrics.
    $\text{AUROC}_{\text{correct~vs.~incorrect}}$ evaluates failure prediction on ID data only, distinguishing between correctly and incorrectly classified samples. 
    The remaining metrics assess OOD detection, either across all ID samples, only correctly classified ones, or only misclassified ones. 
    Each boxplot shows the distribution over \nmodels{} models and \noodcategories{} OOD categories.
    }
    \label{fig:auroc_boxplots_per_method}
\end{figure*}

\paragraph{Are OOD detectors merely identifying misclassified samples?}
We revisit the claim that post-hoc detectors succeed largely because they separate correctly classified ID samples from OOD inputs.
\cref{fig:incorrect_correct_auroc} confirms the strong positive correlation between OOD performance on correctly versus incorrectly classified ID data (Spearman's $\rho = 0.88$, $p \ll 0.001$). 
It also makes the consistent performance gap visually apparent, as nearly all points lie below the $x=y$ identity line, showing that performance is systematically higher on correctly classified samples. 
A notable exception are models trained with MixUp or CutMix, where points for all detectors lie on (or very close to) the identity line, indicating similar OOD performance when conditioning on correct vs. incorrect ID predictions. 
However, the magnitude of this performance gap is highly method-dependent, as detailed in \cref{fig:correct_vs_incorrect_auroc_per_method} and \cref{fig:auroc_boxplots_per_method}.

Classification-based methods like MSP, which are highly sensitive to classification correctness (\ie, high $\text{AUROC}_{\text{correct~vs.~incorrect}}$), exhibit a large performance drop when evaluating on misclassified samples, since their scores are tightly coupled to prediction confidence, these methods risk confusing hard ID examples with true OOD data.
In contrast, methods that leverage richer feature-space representations, like NNGuide and GRAM, show almost no performance gap.
Their near-chance failure detection performance ($\text{AUROC}_{\text{correct~vs.~incorrect}} \approx 50\%$) implies their OOD scoring is largely decoupled from the correctness of the ID classification.

\begin{figure*}[ht]
    \centering
    \begin{subfigure}[b]{\linewidth}
        \centering
        \includegraphics[width=\linewidth]{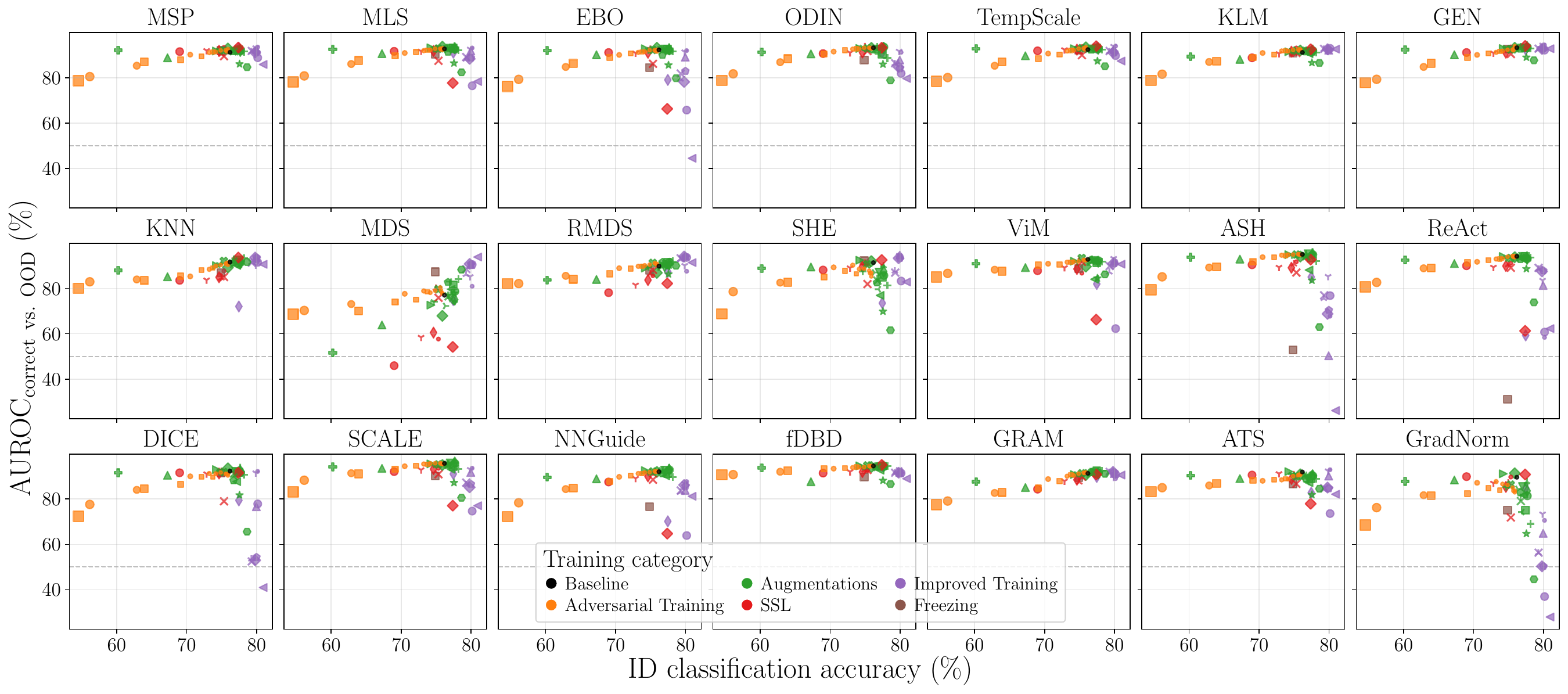}
        \caption{}
    \end{subfigure}

    \vspace{0.5em}
    
    \begin{subfigure}[b]{\linewidth}
        \centering
        \includegraphics[width=\linewidth]{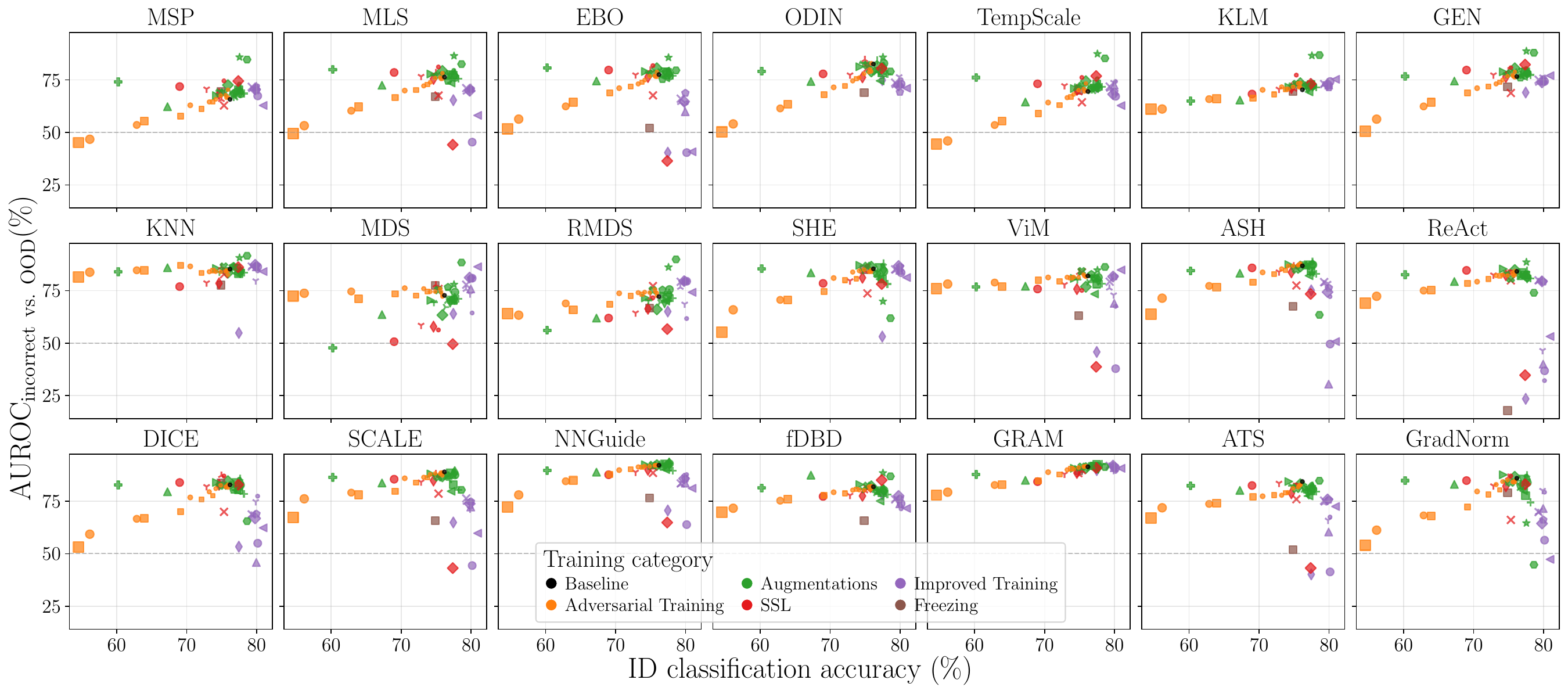}
        \caption{}
    \end{subfigure}
    
    \caption{
    Relation between ID classification accuracy and OOD detection performance.
    Subfigure (a) shows the AUROC for distinguishing correctly classified ID samples from OOD samples, while (b) focuses on incorrectly classified ID samples.
    Each point represents one of the \nmodels{} ResNet-50 models, averaged over \noodsets{} OOD datasets.
    Color indicates the model's training category, while the marker shape uniquely identifies each model within that category.
    }
    \label{fig:correct_acc_incorrect_acc_per_method}
\end{figure*}

\begin{figure*}[ht]
    \centering
    \begin{subfigure}[b]{0.3\linewidth}
        \centering
        \includegraphics[width=\linewidth]{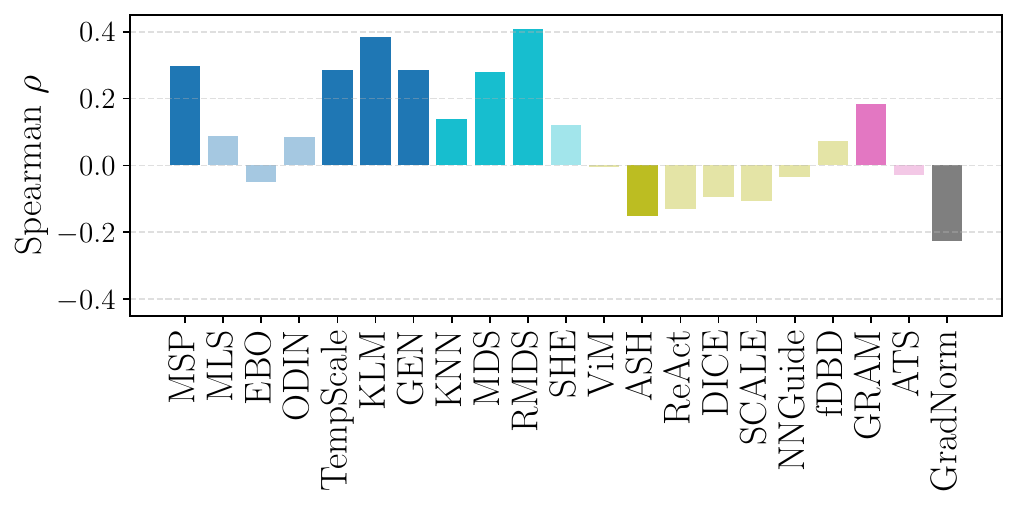}
        \caption{
        Correlation between ID classification accuracy and AUROC.
        }
    \end{subfigure}
    \hfill
    \begin{subfigure}[b]{0.3\linewidth}
        \centering
        \includegraphics[width=\linewidth]{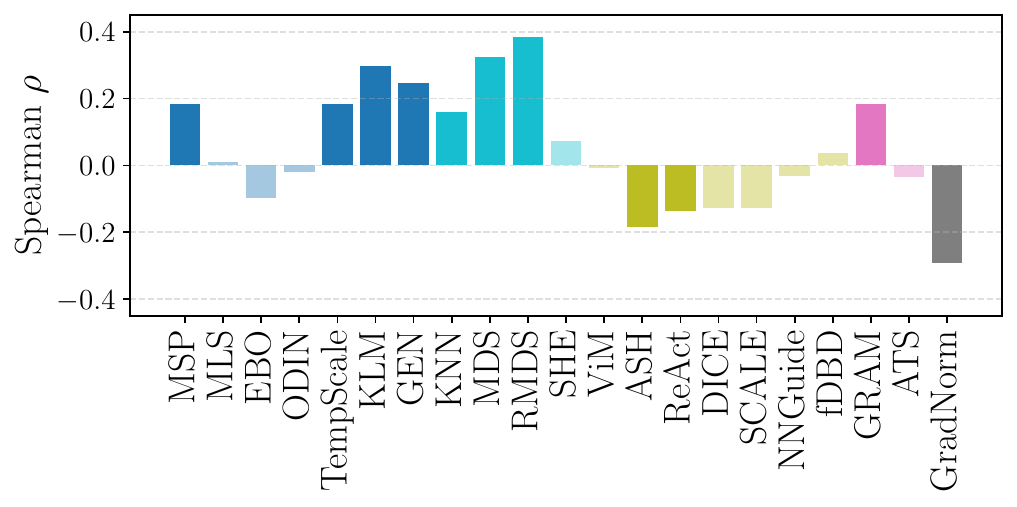}
        \caption{
        Correlation between ID classification accuracy and $\text{AUROC}_{\text{correct~vs.~OOD}}$.
        }
    \end{subfigure}
    \hfill
    \begin{subfigure}[b]{0.3\linewidth}
        \centering
        \includegraphics[width=\linewidth]{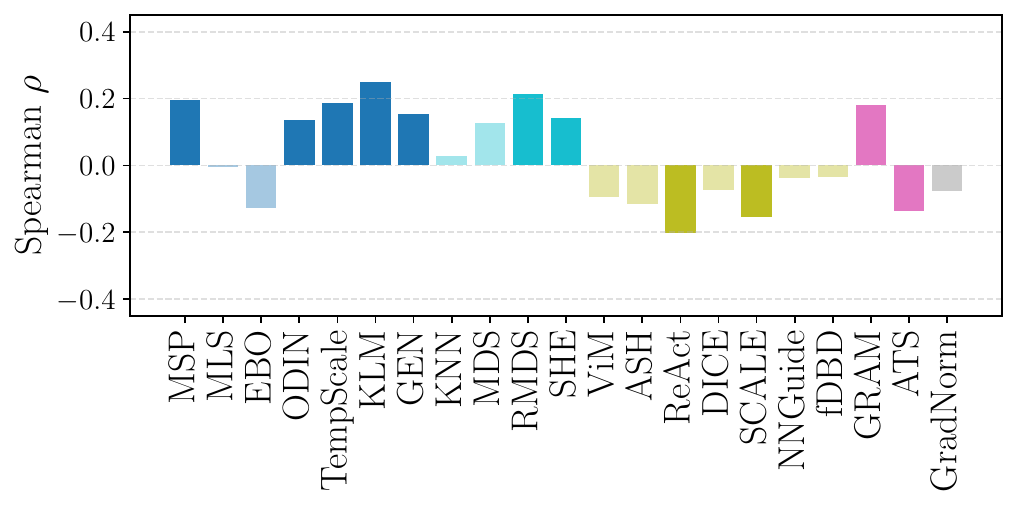}
        \caption{
        Correlation between ID classification accuracy and $\text{AUROC}_{\text{incorrect~vs.~OOD}}$.
        }
    \end{subfigure}
    
    \caption{
    Relationship between ID classification accuracy and OOD detection performance.
    Spearman rank correlation~($\rho$) between ID classification accuracy and OOD-detection AUROC for each detector: (a) all ID samples, (b) only correctly classified ID samples, and (c) only misclassified ID samples.
    Bars are sorted and color-coded according to the method's OOD detection category (\mycircle{classification} classification-based, \mycircle{feature} feature-based,  \mycircle{hybrid} hybrid, \mycircle{inter} intermediate-feature, \mycircle{gradients} gradients).
    Non-significant correlations ($p \geq 0.05$) are shown with reduced opacity.
    Statistics are computed over \nmodels models and \noodcategories OOD categories.   
    }
    \label{fig:spearman_correct_incorrect_ood_acc}
\end{figure*}

In \cref{fig:spearman_correct_incorrect_ood_acc,fig:correct_acc_incorrect_acc_per_method}, we correlate the OOD detection performance (AUROC) with the ID classification accuracy. 
This analysis is performed for all ID samples, and we further dissect the behavior by also considering the subsets of correctly and incorrectly classified samples separately (\cref{fig:correct_acc_incorrect_acc_per_method}). 
The Spearman correlation coefficients (\cref{fig:spearman_correct_incorrect_ood_acc}) reveal a consistently weak or statistically non-significant relationship across all three groups (\ie, all, correct, and incorrect), echoing the main manuscript finding.
This result diverges from prior work~\cite{Humblot-Renaux_2024_CVPR}; while they performed a similar analysis, they observed a strong overall correlation that was almost entirely driven by the performance on correctly classified ID samples ($\text{AUROC}_{\text{correct~vs.~OOD}}$)

While $\text{AUROC}_{\text{incorrect~vs.~OOD}}$ can approach random chance for some model--method pairs, this is not the case for well-matched configurations (\cref{fig:correct_acc_incorrect_acc_per_method}). 
For the baseline model---representing the default benchmark setting where the model-method fit is strong---every single detector performs significantly better than random guessing.
This demonstrates a genuine ability to distinguish true OOD samples from a model's own most challenging ID examples, proving that---while misclassifications impair performance---these methods are fundamentally more than mere failure detectors.

\paragraph{Where does the AUROC variance come from?}
\cref{tab:anova_effects} lists the complete F-values, p-values, and variance shares of the three-way ANOVA; all main effects and interactions are significant ($p \ll 0.001$).
To rule out a single OOD category artifact, we reran the ANOVA four times, each time omitting one OOD category.
\cref{tab:anova_leaveoneout} shows the variance shares.
Leaving out the hardest split (near-OOD) drops the OOD category main effect to $6.59\%$, but the \textit{model} $\times$ \textit{method} interaction increases to $33.36\%$, revealing model-detector coupling that had been masked by uniformly low AUROC on the toughest OOD category.
When far- or extreme-OOD is omitted, the OOD-category term remains dominant ($\approx 40\%$) while the interaction never falls below $16\%$. 
The residual variance is stable across all runs.
Thus, no single dataset dictates the conclusions; indeed, model–OOD method compatibility becomes more salient once the most challenging category is removed, underscoring the need for a diverse OOD benchmark.
\begin{table}
\centering\footnotesize
\resizebox{\linewidth}{!}{
\begin{tabular}{lrrrr}
\toprule
\textbf{Effect} & \textbf{F-value} & \textbf{p-value} & \textbf{Variance Share (\%)} \\
\midrule
Method & 156.69 & $\ll 0.001$ & 7.08 \\ 
Model & 77.90 & $\ll 0.001$ & 9.69 \\ 
OOD Category & 5045.22 & $\ll 0.001$ & 34.22 \\ 
Model $\times$ Method & 8.47 & $\ll 0.001$ & 21.05 \\ 
Model $\times$ OOD Category & 9.10 & $\ll 0.001$ & 3.39 \\ 
Method $\times$ OOD Category & 39.05 & $\ll 0.001$ & 5.30 \\ 
Model $\times$ Method $\times$ OOD Group & 1.16 & $\ll 0.001$ & 8.66 \\ 
Residual & --- & --- & 10.62 \\ 
\bottomrule
\end{tabular}
}
\caption{
Three-way ANOVA decomposition of AUROC variance across models, OOD detection methods, and OOD dataset categories. 
The table reports the F-statistic, significance level (p-value), and proportion of explained variance for each main effect and interaction. 
}
\label{tab:anova_effects}
\end{table}

\begin{table*}
\centering\footnotesize
\resizebox{\textwidth}{!}{
\begin{tabular}{lcccccccc}
\toprule
\textbf{Left-Out} & \textbf{Model} & \textbf{Method} & \textbf{OOD Category} & \textbf{Model$\times$Method} & \textbf{Model$\times$OOD Category} & \textbf{Method$\times$OOD Category} & \textbf{3-Way Interaction} & \textbf{Residual} \\
\midrule
Near      & 16.66  & 10.02 &  \phantom{0}6.78 & 33.87 & 4.33 & 6.41 & 11.43  & 10.55 \\
Far       &  \phantom{0}9.59  &  \phantom{0}5.16 & 42.15 & 16.51 & 3.44 & 6.25 &  \phantom{0}8.45  &  \phantom{0}8.44 \\
Extreme   &  \phantom{0}8.41  &  \phantom{0}8.62 & 40.23 & 19.79 & 2.17 & 2.60 &  \phantom{0}6.46  & 11.67 \\
Synthetic & 11.49  &  \phantom{0}8.04 & 31.49 & 22.49 & 2.76 & 4.63 &  \phantom{0}6.39  & 12.77 \\
\bottomrule
\end{tabular}
}
\caption{
Explained variance from leave-one-out 3-way ANOVA (factors: model, method, OOD category). 
Each row excludes one OOD group and recomputes variance proportions.
All reported values are in percentage and statistically significant ($p \ll 0.001$).
}
\label{tab:anova_leaveoneout}
\end{table*}

\begin{figure*}[t]
    \centering
    \includegraphics[width=\linewidth]{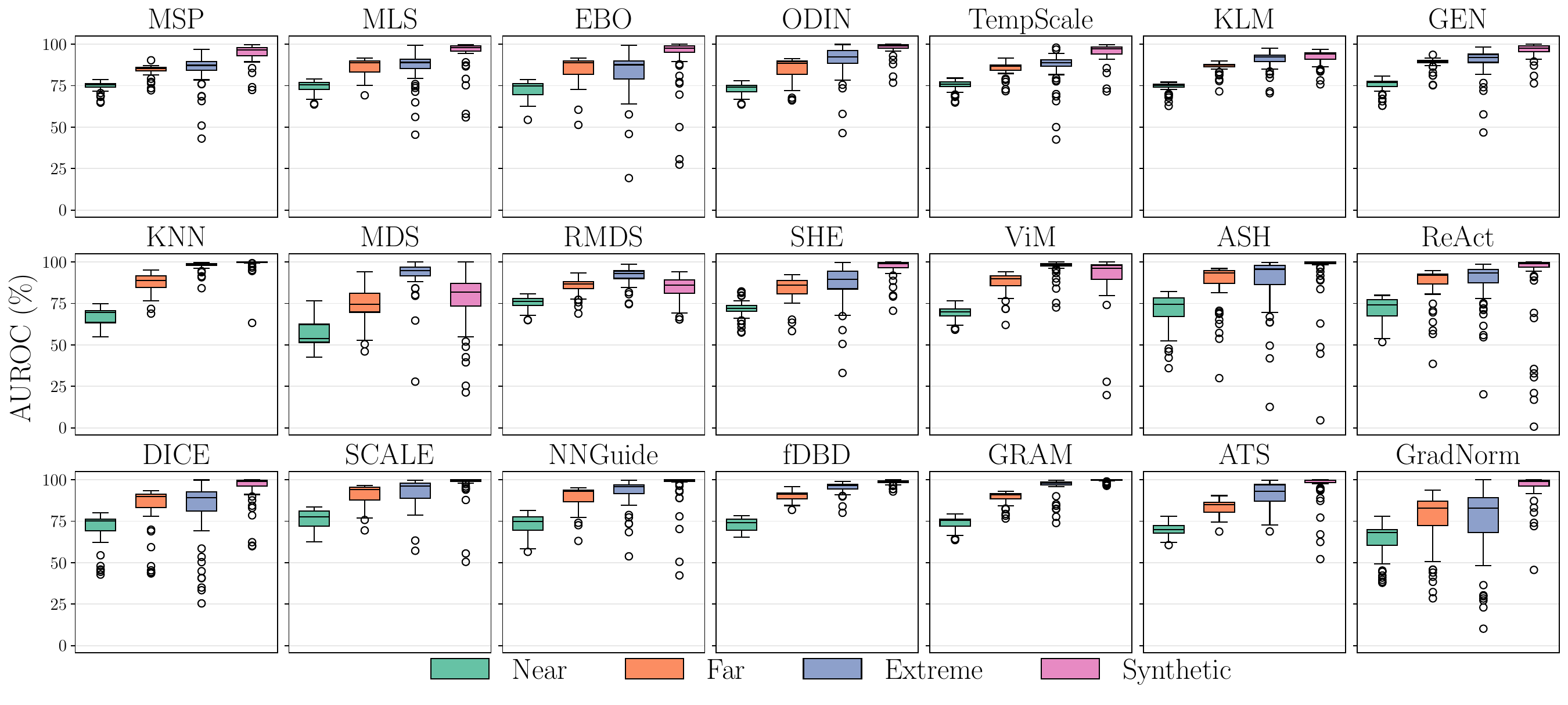}
    \caption{
    OOD detection performance across different OOD categories (near, far, extreme, and synthetic).
    Each boxplot shows the distribution over \nmodels models.
    }
    \label{fig:auroc_per_ood_group_per_method}
\end{figure*}

\paragraph{How robust are detection methods across training variants?}
The robustness of OOD detection methods also depends on the nature of the distributional shift. 
\cref{fig:auroc_per_ood_group_per_method} shows the OOD detection performance for each method across the four OOD categories, revealing several key insights.

First, as expected, performance is generally lowest for the most challenging near-OOD datasets, where the semantic similarity with the ID data is highest. 
Most methods struggle to achieve high AUROC scores in this setting, confirming the difficulty of this benchmark.

Second, and more surprisingly, the variance in performance across our \nmodels models is often highest for the supposedly easier extreme- and synthetic-OOD categories. 
This suggests that the choice of training strategy can have a more pronounced and unpredictable impact on a method's effectiveness when the domain shift is large but structurally simple (\eg, ImageNet vs. MNIST).

This highlights a critical aspect of robustness: a method that appears stable and effective on near-OOD data may become unreliable on other types of shifts, and vice versa. 
For example, the high variance of some model enhancement methods on extreme- and synthetic-OOD data may not just stem from a sensitivity to low-level statistics, but also from operating on final-layer features where discriminative information for structurally simple OOD data might be diminished. 
This hypothesis is supported by prior work~\cite{Krumpl2024_ats}, which showed that simpler OOD tasks are often more easily solved in a model's earlier layers. 
The notable robustness of GRAM, which leverages intermediate features, on these same categories lends further support to this idea, suggesting that access to earlier representations is key for handling such shifts. This underscores the necessity of benchmarking on a wide range of OOD test sets to gain a complete picture of a method's generalization capabilities.

\begin{figure}[t]
  \centering
   \includegraphics[width=1.0\linewidth]{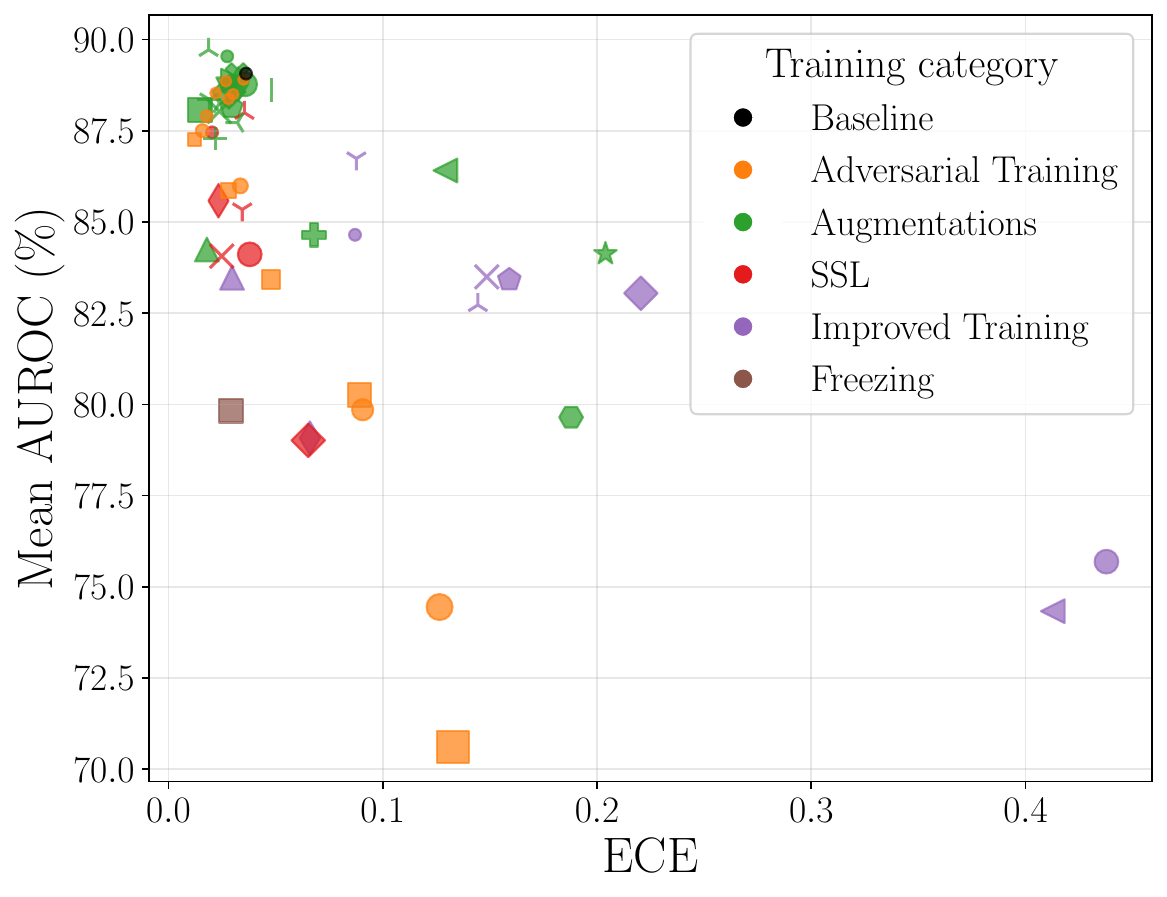}

   \caption{
        Relationship between the expected calibration error (ECE) and the OOD detection performance.   
        Each point represents one of \nmodels models, with performance averaged across all \nmethods OOD detection methods and \noodsets OOD datasets.
   }
   \label{fig:auroc_vs_ece}
\end{figure}

\paragraph{Relationship with Model Calibration}
To investigate if other ID metrics are better predictors of OOD performance than accuracy, we analyzed the relationship between Expected Calibration Error (ECE) and the OOD detection performance (\cref{fig:auroc_vs_ece}). 
Globally, we observe a weak negative correlation (Spearman's $\rho = -0.17, p \ll 0.001$), which, while more consistent than the correlation with ID classification accuracy ($\rho = 0.04$), remains a poor proxy for OOD detection performance.

\begin{figure}[t]
  \centering
   \includegraphics[width=1.0\linewidth]{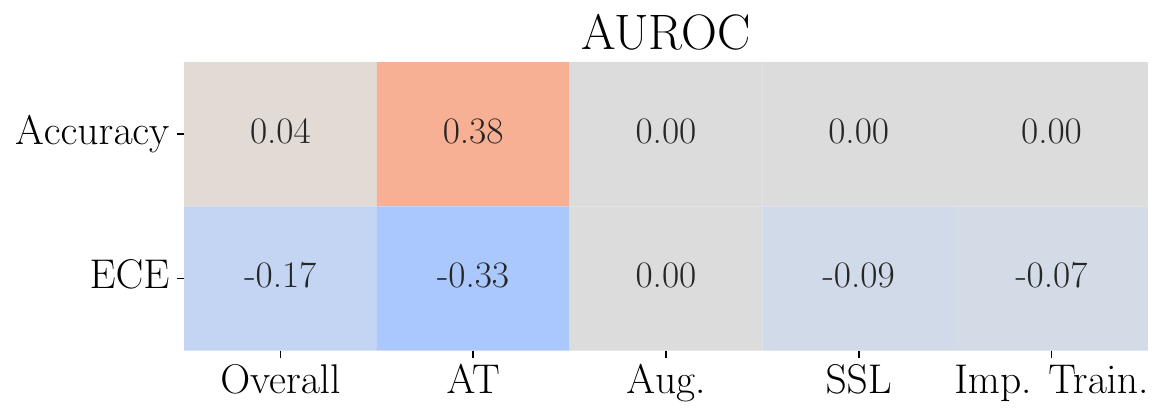}

   \caption{
        Spearman correlation coefficients ($\rho$) between OOD detection performance (AUROC) and in-distribution (ID) performance metrics (accuracy and ECE), computed across all models, OOD methods, and OOD categories (\textit{Overall}), and separately for adversarial training (\textit{AT}), data augmentations (\textit{Aug.}), self-supervised learning (\textit{SSL}), and improved training recipes (\textit{Imp. Train.}). 
        Non-significant correlations ($p \geq 0.05$) are set to 0. Note that Spearman $r$ reflects monotonic relationships and may not capture non-monotonic trends.
   }
   \label{fig:spearman_overall_train_category}
\end{figure}

A breakdown by training category (\cref{fig:spearman_overall_train_category}) reveals that this global correlation is a misleading artifact.
The trend is driven almost entirely by the adversarial training regime ($\rho = -0.33$). 
At the same time, models trained with augmentations, SSL, or improved recipes show little to no correlation between their calibration and OOD detection performance. 

This finding underscores that OOD detection performance is too complex to be reliably predicted by a single ID metric, such as accuracy or calibration. 
While correlations may appear within specific subgroups (\eg, training strategies or OOD detection methods), such as adversarially trained models, they do not imply causality and fail to generalize across the diverse landscape of training strategies, making them unreliable as universal proxies.

\begin{figure}[t]
  \centering
   \includegraphics[width=1.0\linewidth]{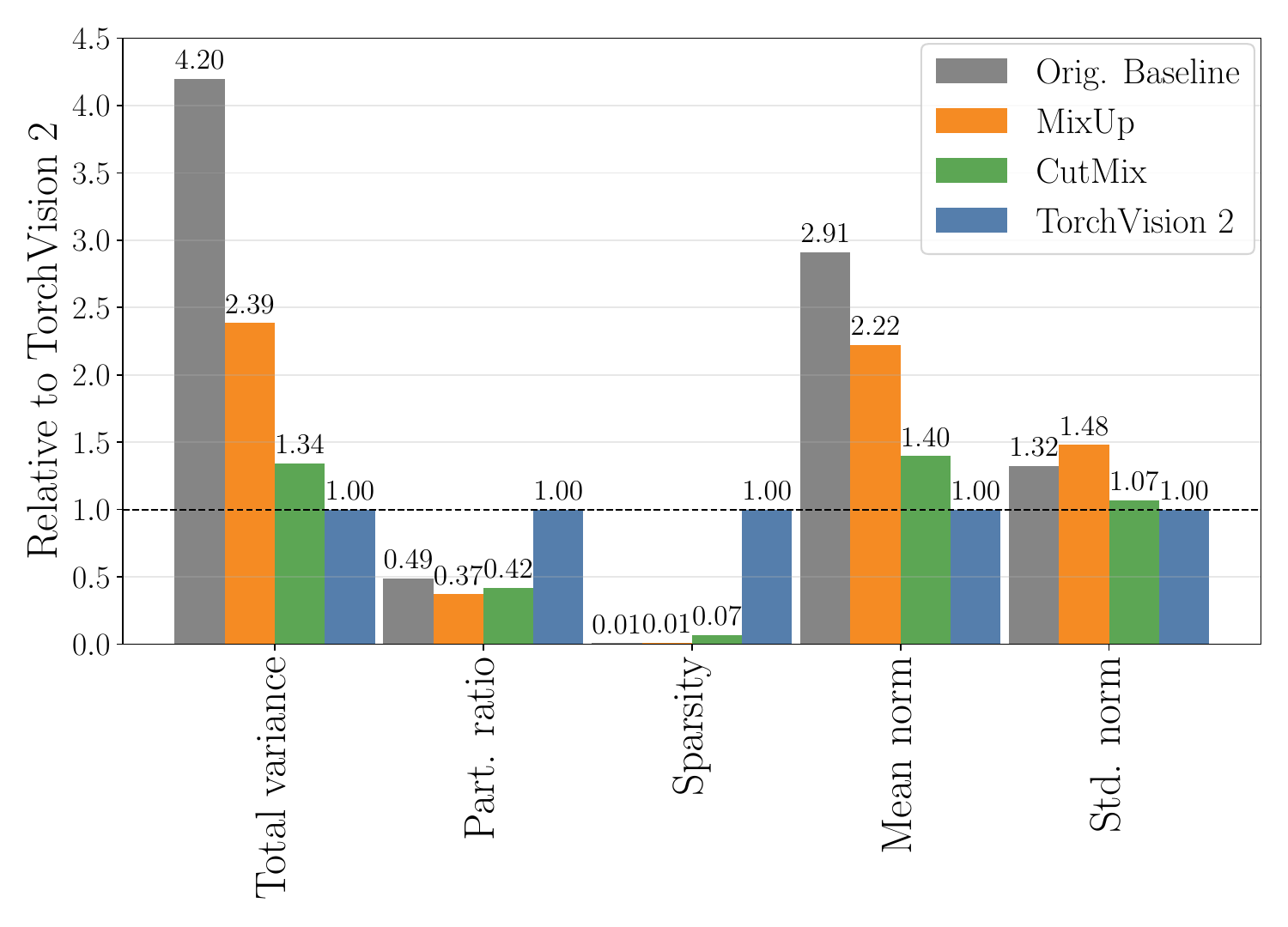}

   \caption{
    Feature-space metrics for ResNet-50 baseline, MixUp, CutMix, and TorchVision 2 on the ImageNet test set, computed from penultimate-layer embeddings and shown relative to TorchVision 2 (set to 1.0).
    We report five complementary statistics: i) total variance, the trace of the covariance matrix measuring overall spread of embeddings; ii) participation ratio, the effective dimensionality of the feature space; iii) sparsity, the fraction of activations below $10^{-3}$; iv) mean feature norm, the average $l_2$-norm of embedding vectors; and v) standard deviation of feature norms, capturing variability in embedding magnitudes.
   }
   \label{fig:penult_feature_metrics}
\end{figure}

\paragraph{Feature-Space Analysis and Robustness of OOD Detection Methods.}
To better understand why advanced training recipes degrade OOD detection, we analyze feature-space statistics for four ResNet-50 models: the baseline~\cite{He2015_resnet}, MixUp~\cite{zhang2018mixup}, CutMix~\cite{Sangdoo2019_cutmix}, and the TorchVision 2 recipe~\cite{Paszke19_pytorch, Vasilis2023_TV2} (which includes MixUp, CutMix together with additional regularization such as label smoothing, stronger augmentation, and EMA; see \cref{fig:penult_feature_metrics}).
We report five complementary metrics: total variance (spread of embeddings), participation ratio (effective dimensionality), sparsity (fraction of near-zero activations), and the mean and standard deviation of feature norms.

While MixUp, CutMix, and TorchVision 2 achieve higher ID accuracy than the baseline, their internal representations become progressively more compressed.
From the baseline through MixUp and CutMix to TorchVision 2, we observe a clear progression.
MixUp reduces variance and feature norms while lowering the participation ratio, suggesting a lower-rank embedding. 
CutMix shows similar but slightly stronger effects, with variance/norm reduced further and sparsity moderately increased. 
TorchVision 2 amplifies these trends: variance and norms collapse, sparsity increases by more than two orders of magnitude, and the representation is flattened. 
Thus, while all three advanced recipes achieve higher ID accuracy than the baseline, they also progressively compress and sparsify the embedding space.

\begin{figure*}
    \centering
    \includegraphics[width=\linewidth]{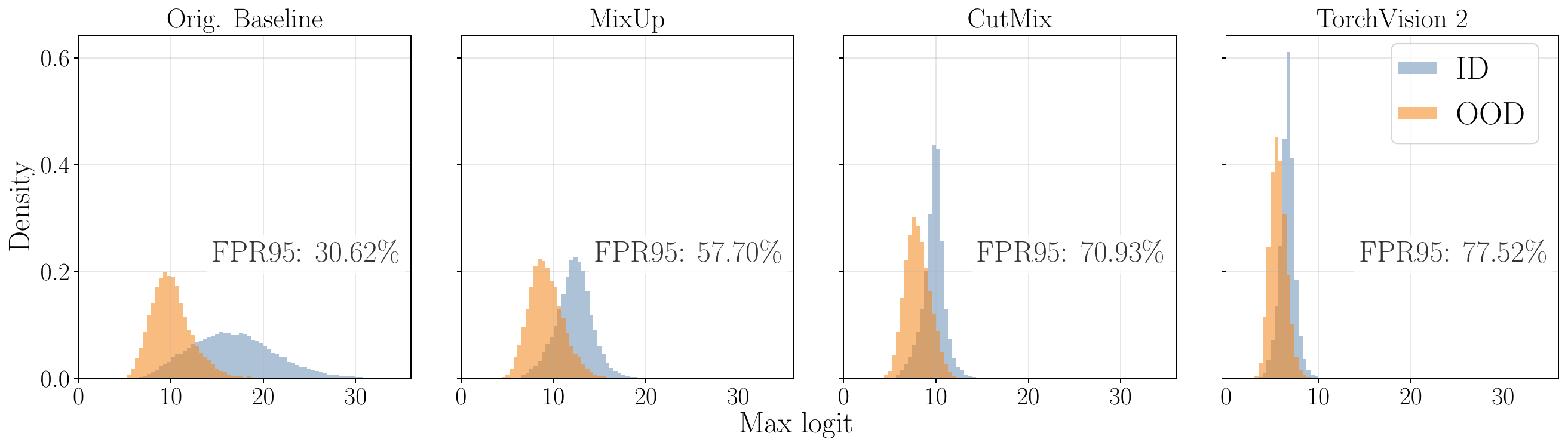}
    \caption{
    Max-logit distributions for ResNet-50 baseline, MixUp, CutMix, and TorchVision 2 with ImageNet (ID) and iNaturalist (OOD). 
    The plots show the distribution of the maximum predicted logit for ID and OOD samples, together with the corresponding FPR95 values.
    }
    \label{fig:max_logit}
\end{figure*}

\begin{figure*}
    \centering
    \includegraphics[width=\linewidth]{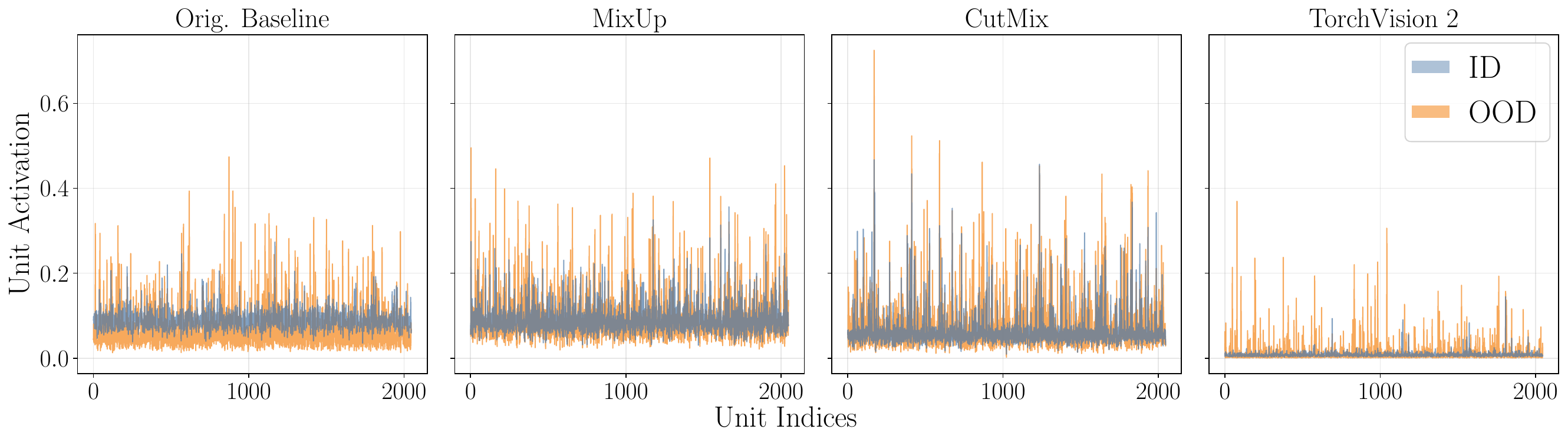}
    \caption{
    Distribution of per-unit activations in the penultimate layer for ImageNet (ID) and iNaturalist (OOD) across ResNet-50 baseline, MixUp, CutMix, and TorchVision 2.
    }
    \label{fig:penult_profile}
\end{figure*}

These shifts are also mirrored in the logit and embedding space (see \cref{fig:max_logit,fig:penult_profile}). 
The max-logit distributions become narrower and show increasing ID-OOD overlap: baseline leaves a clear margin (FPR95 = 30.62\%), MixUp reduces separation (57.70\%), CutMix worsens it further (70.93\%), and TorchVision 2 nearly eliminates it (77.52\%). 
Likewise, the penultimate-layer activation distributions show that the characteristic pattern described by Sun \etal~\cite{sun2021_ood_react}---a near-constant mean activation for ID samples and lower but more variable activations for OOD samples, which ReAct exploits via activation clipping---progressively changes under MixUp, CutMix, and TorchVision 2.
As a result, activation-shaping detectors such as ReAct---whose efficacy depends on clipping high activations---lose discriminative power, reflected in a significant performance drop: FPR95 increases from 16.75\% (baseline) to 40.46\% (MixUp), 58.51\% (CutMix), and 88.55\% (TorchVision 2).

In contrast, feature-based methods (\eg, KNN, GRAM, RMD) that leverage distances or higher-order statistics rather than specific activation characteristics, and therefore remain comparatively robust under increasing regularization. 
Altogether, these results show that although MixUp, CutMix, and TorchVision 2 improve ID accuracy, they also systematically reshape the feature space in ways that disadvantage activation-based detectors while leaving geometry-based or magnitude-agnostic approaches more stable. 
This provides further evidence for our central finding that improvements in ID accuracy do not necessarily yield better OOD detection, underscoring the strong dependency between the underlying model and the effectiveness of a given OOD detection method.

\section{Results on ViT}
\label{sec:supp:result_vit}

To test whether our findings extend beyond ResNets, we also evaluate Vision Transformer (ViT-B/16) models~\cite{dosovitskiy2021vit}, that originate from AugReg~\cite{steiner2022_vit_augreg}, Masked Autoencoders (MAE)~\cite{He_2022_MAE}, Data-Efficient Image Transformers (DeiT)~\cite{touvron21_deit}, and Sharpness-Aware Minimization (SAM)~\cite{chen2022_vit_sam}. 
As with ResNet, all models are trained exclusively on the ILSVRC2012 subset of ImageNet to prevent OOD contamination.

\begin{figure}[t]
  \centering
   \includegraphics[width=1.0\linewidth]{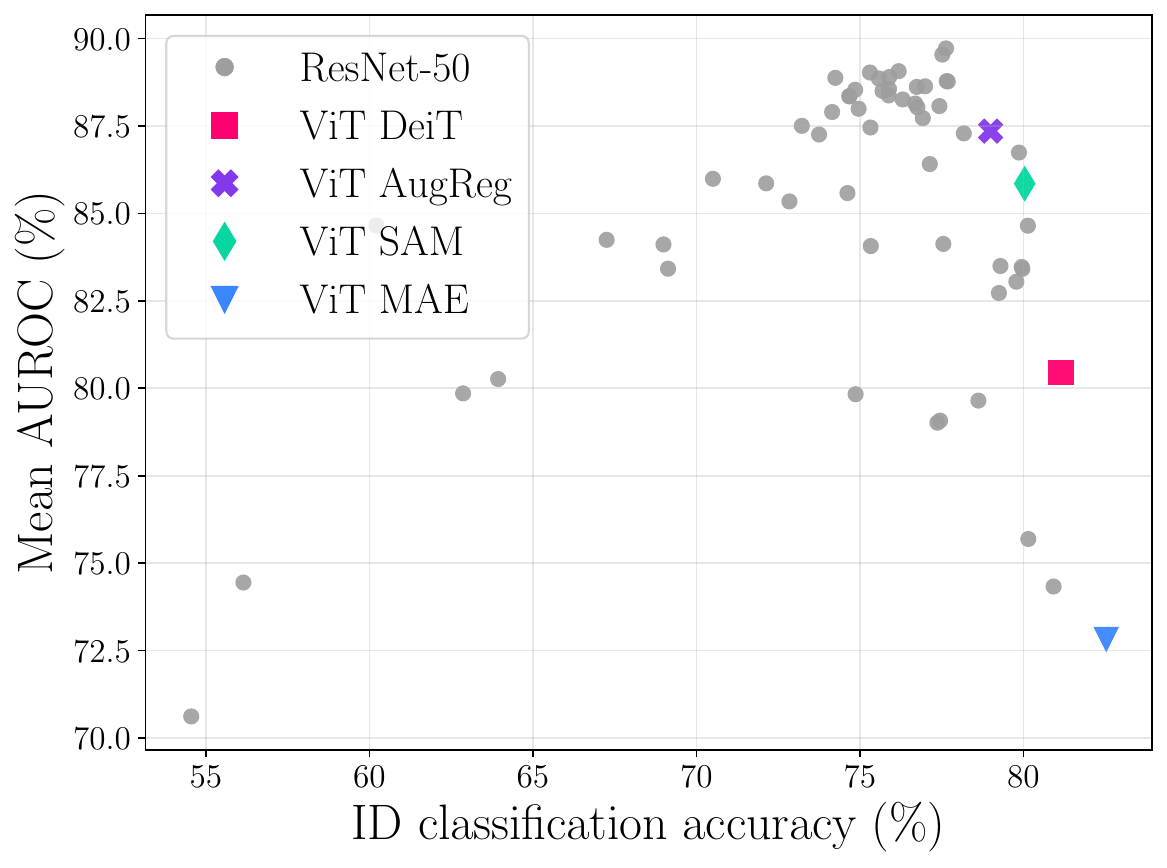}

   \caption{
    Relationship between ID accuracy and OOD detection performance (AUROC) for \nmodels ResNet-50 and four ViT-B/16 models. 
    Each point corresponds to a specific training strategy on ImageNet (ID), with OOD performance averaged over \nmethods detection methods and \noodsets OOD datasets.
   }
   \label{fig:auc_vs_acc_resnet_vit}
\end{figure}

\begin{figure*}
    \centering
    \includegraphics[width=\linewidth]{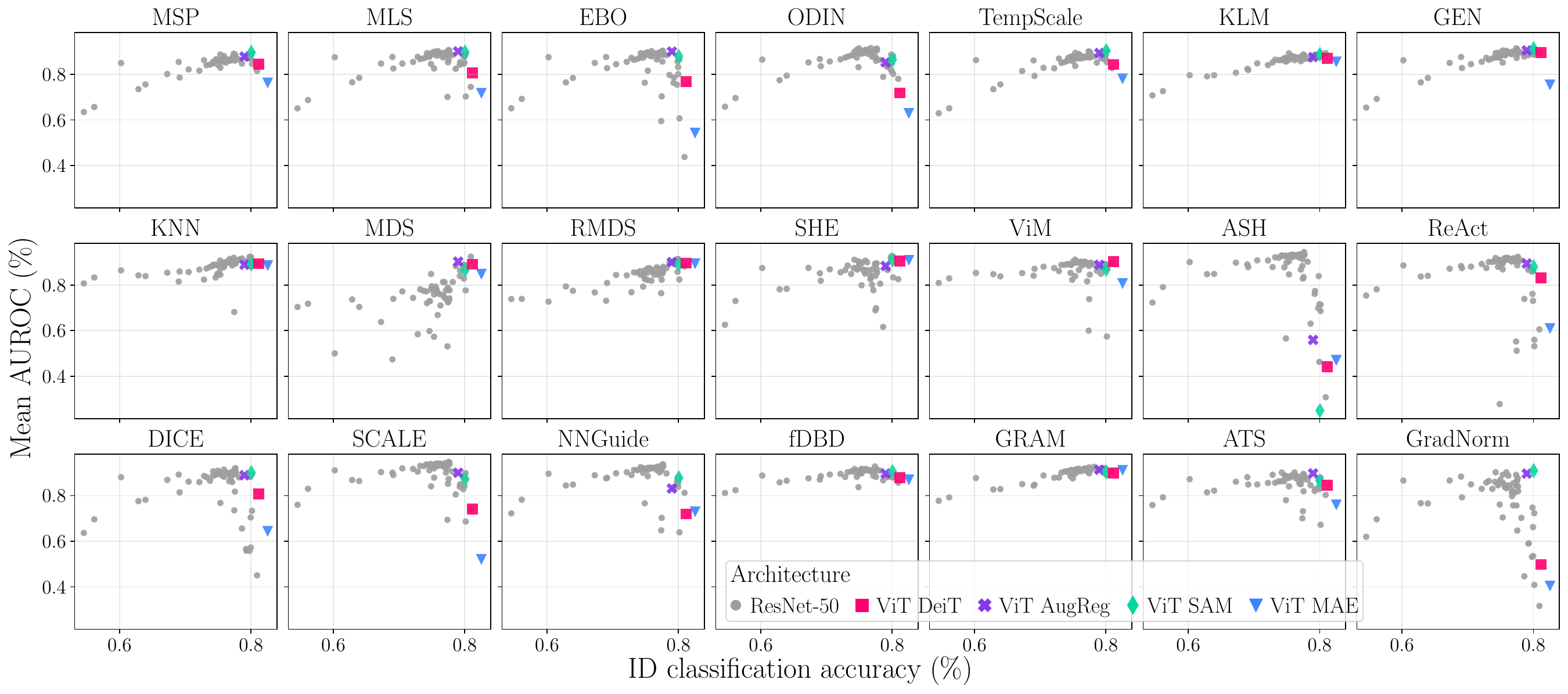}
    \caption{
    Relationship between ID accuracy and OOD detection performance (AUROC) for ResNet-50 and ViT-B/16 across individual OOD detection methods.
    Each panel shows one OOD detection method, with points corresponding to different training strategies (\nmethods ResNet-50 and four ViT-B/16 models trained on ImageNet).
    OOD performance is averaged across \noodsets OOD datasets.
    }
    \label{fig:auroc_vs_acc_per_ood_method_resnet_vit}
\end{figure*}

Consistent with our ResNet results, ViTs achieve higher ID accuracy but do not exhibit improved OOD detection performance (see \cref{fig:auc_vs_acc_resnet_vit}). 
At the OOD detection method level (\cref{fig:auroc_vs_acc_per_ood_method_resnet_vit}), we again observe a clear dichotomy: feature-based methods that rely on distances or higher-order statistics (\eg, KNN, RMDS, GRAM) remain comparatively robust, while model-enhancement methods that depend on shaping specific activation patterns degrade substantially.

These findings reinforce our central claim that better ID accuracy does not guarantee better OOD detection, even for more modern, higher-capacity architectures. 
They also support recent evidence~\cite{zhang2024openood} that many OOD detection methods have been implicitly tuned to CNN-style representations, and may overfit to the activation characteristics of ResNets rather than transfer robustly to other architectures.

\end{document}